\documentclass[conference]{IEEEtran}
\IEEEoverridecommandlockouts
\usepackage{cite}
\usepackage[size=smaller]{caption}
\usepackage{amsmath,amssymb,amsfonts, bbm}
\usepackage{subcaption}
\usepackage{stfloats} 
\usepackage{booktabs} 
\usepackage{graphicx}
\usepackage{textcomp}
\usepackage{xcolor}
\usepackage{array}

\usepackage[ruled,vlined]{algorithm2e}
\usepackage{multicol}

\def\BibTeX{{\rm B\kern-.05em{\sc i\kern-.025em b}\kern-.08em
    T\kern-.1667em\lower.7ex\hbox{E}\kern-.125emX}}
\begin{document}

\title{Input-Triggered Hardware Trojan Attack on \\Spiking Neural Networks\thanks{This work was supported by the French National Research Agency (ANR) and the UK Research and Innovation (UKRI), Engineering and Physical Sciences Research Council (EPSRC), through the European CHIST-ERA program under the project TruBrain (Grants N$^{\mbox{\scriptsize o}}$ ANR-23-CHR4-0004-01 and EP/Y03631X/1, respectively) and by the European Network of Excellence dAIEDGE (Grant  N$^{\mbox{\scriptsize o}}$ 101120726).}
}


\author{\IEEEauthorblockN{Spyridon Raptis\IEEEauthorrefmark{1}, Paul Kling\IEEEauthorrefmark{1}, Ioannis Kaskampas\IEEEauthorrefmark{1}, Ihsen Alouani\IEEEauthorrefmark{2}, Haralampos-G. Stratigopoulos\IEEEauthorrefmark{1}\\} \IEEEauthorblockA{\IEEEauthorrefmark{1}\small{Sorbonne Universit\'{e}, CNRS, LIP6, Paris, France}\\}\IEEEauthorblockA{\IEEEauthorrefmark{2}\small{CSIT, Queen’s University Belfast, Belfast, UK}}}

\maketitle

\begin{abstract}

Neuromorphic computing based on spiking neural networks (SNNs) is emerging as a promising alternative to traditional artificial neural networks (ANNs), offering unique advantages in terms of low power consumption. However, the security aspect of SNNs is under-explored compared to their ANN counterparts. As the increasing reliance on AI systems
comes with unique security risks and challenges, understanding the vulnerabilities and threat landscape is essential as neuromorphic computing matures. 
In this effort, 
we propose a novel input-triggered Hardware Trojan (HT) attack for SNNs. 
The HT mechanism is condensed in the area of one neuron. The trigger mechanism is an input message crafted in the spiking domain such that a selected neuron produces a malicious spike train that is not met in normal settings. This spike train triggers a malicious modification in the neuron that forces it to saturate, firing permanently and failing to recover to its resting state even when the input activity stops. The excessive spikes pollute the network and produce misleading decisions. We propose a methodology to select an appropriate neuron and to generate the input pattern that triggers the HT payload. The attack is illustrated by simulation on three popular benchmarks in the neuromorphic community. We also propose a hardware implementation for an analog spiking neuron and a digital SNN accelerator, demonstrating that the HT has a negligible area and power footprint and, thereby, can easily evade detection.

\end{abstract}


\section{Introduction} \label{sec:introduction}

Neuromorphic architectures having as basis spiking neural networks (SNNs) offer a fundamentally different approach in information processing compared to conventional artificial neural networks (ANNs). By mimicking the functionality of the biological brain, SNNs process data in an asynchronous, event-driven fashion. This property makes SNNs less computationally and energy-intensive, thus carrying promising opportunities for the increasingly demanding requirements of artificial intelligence (AI) \cite{RoJaPa19, SKPMDK22}. In particular, in SNNs the information is processed in the form of spike trains and is encoded in the timing between spikes or in the spike firing rate. Spikes are processed as soon as they are generated, thus offering real-time processing and a low-latency inference. Power is consumed only when a neuron fires a spike, allowing SNNs to achieve ultra low-power computations \cite{RoJaPa19, SKPMDK22}.
Nowadays, there are intensive efforts in designing hardware platforms for neuromorphic computing \cite{FGTP14,Merolla14,Loihi18,Camunas18,SFMRWQ22}.


With AI applications going mainstream, including in safety-critical and security-sensitive domains, there has been extensive interest in the security of the different architectures at different stages of their training and deployment pipeline \cite{QLZH22,WYJTC22, CaWa17}.
The threat landscape includes: (a) misusing the AI system for malicious purposes, i.e., via adversarial \cite{SZSBEG14} and backdoor attacks \cite{GLD-GG19}; (b) stealing or reverse engineering the neural network model which is often considered an asset, for example via a power or timing side-channel attack \cite{HuZhSu18}; (c) compromising the functionality of the AI application, i.e., via fault injection attacks \cite{RaHeFa19} or Hardware Trojans (HTs) \cite{ClLa19}; (d) and undermining the privacy, i.e., inference attacks \cite{CTWJH-VLRBSEOR21} aiming at deducing sensitive information about the data or the model use. Security in AI should be a priority so as to ensure its safe integration into society, prevent misuse, and foster trust in this transformative technology.
Specifically for SNNs, security threats that have been studied include adversarial attacks \cite{BaSiRa18,BLHSS18,SPSLPR19,E-AMSA21,MPMMS21,LHDWLDLX23,MNKHMS20,NSHM22}, backdoor attacks \cite{AEPU24}, fault injection attacks \cite{NLEKKG22}, side-channel attacks \cite{GARHTS21,NRTKG23,GoDaSu24}, and HTs \cite{VMAMS20}.

In this work, we focus on the HT attack. A HT is a malicious modification of the hardware consisting of a triggering mechanism and a payload mechanism, i.e., the malicious functionality executed when the HT is triggered \cite{BhTe18}. A plethora of HT designs has been proposed to date for different circuit classes that can be generic or specific within a circuit class. Consequently, a taxonomy is proposed according to the attacker, victim circuit type, attack insertion level (i.e., RTL, gate-level, transistor-level, layout level), trigger type (i.e., random, input-defined, always on), and payload (i.e., information leakage, performance degradation, function modification, denial-of-service) \cite{TeKo10, KRRT10}. An attacker aims at inserting a stealthy HT that evades known detection countermeasures, whereas a defender aims at preventing HT insertion or detecting the HT presence. HT designs range from simple ones, i.e., combinatorial and sequential \cite{BHBN14}, to more sophisticated ones, i.e., hidden side-channels \cite{LKGPB09}, silicon wearout mechanisms \cite{SWRPWC10}, changing dopant polarity in active areas of transistors \cite{BRPB14}, siphoning charge from victim wires known as A2 attack \cite{YHDAS16}, digital-to-analog attacks \cite{EDiNSPLAS21}, leaking sensitive data through covert communication channels \cite{D-RAAS24}, etc.

In this work, we propose a  novel SNN-specific HT attack where the payload is triggered through an input pattern. The proposed HT is condensed in the area within a single spiking neuron. The triggering mechanism monitors the output of the neuron and activation happens when a specific spike pattern appears. We developed an algorithm that uses gradient back-propagation to craft a specific input in the spiking domain. This input, when propagated through the network layers, generates the desired specific spike pattern at the output of the target neuron. In other words, the HT is \textbf{externally activated} with its triggering being input-referred. The payload mechanism forces the victim neuron to saturate, producing non-stop spikes which spread through the network and propagate to the output to change the network's decision. The saturation is permanent for any following input, resulting in high probability of misprediction, i.e., essentially the activation leads to denial-of-service. In \cite{VMAMS20}, which is the only existing HT proposal for SNNs, the HT is activated when a neuron generates a given high number of spikes and the payload consists of causing bit flips in the on-chip memory storing the synaptic weights. Compared to \cite{VMAMS20}, the proposed HT is localized and, thereby, it is much harder to detect by the defender. Furthermore, in \cite{VMAMS20}, there was no hardware implementation. In this work, we propose hardware implementations for both SNN hardware design paradigms: analog and digital.

The remainder of this paper is structured as follows. In Section \ref{sec:threat_model}, we present the threat model. The attack is described in Section \ref{sec:attack}, including an overview of the working principle, the methodology to select the victim neuron, and the algorithm to generate the input trigger. In Section \ref{sec:results}, we demonstrate simulation results. In Section \ref{sec:HT_design}, we present the hardware implementations and results. In Section \ref{sec:countermeasures}, we discuss possible countermeasures. Section \ref{sec:conclusion} concludes this paper.

\section{Threat model} \label{sec:threat_model}

\textbf{Context and Assumptions.} We consider a scenario where a client owns a proprietary spiking dataset and wants to train an SNN and deploy it on a hardware accelerator for a specific application. 
The client does not have the resources, i.e., GPU cluster, to train the model. The neuromorphic system provider offers cloud access for training as a part of the package to the client in a typical Machine Learning as a Service (MLaaS) setting.
The adversary is the neuromorphic system provider.
We assume the adversary has access to the SNN model architecture and dataset during the outsourced training phase. The SNN hardware accelerator is equipped with a compiler that maps the trained SNN model onto hardware-specific units.

\textbf{Adversary's Capabilities.} The neuromorphic system provider, i.e. the adversary, has the following capabilities:

\noindent\textit{(i) Supply Chain Capability:} Similar to state-of-the-art HT threat model, we assume that during the design phase of the hardware accelerator, the adversary can insert a stealthy HT targeting specific hardware neurons.

\noindent\textit{(ii) Access to Model and Dataset}: By offering cloud-based training in a MLaaS setting, the adversary has access to both the SNN architecture and dataset provided by the client.

\textbf{Adversary's Objectives.} The adversary's main objective is to compromise the integrity of the SNN hardware accelerator and its deployed application for own benefit, or is commissioned to do so by a third-party attacker, without raising suspicion. Specifically, the adversary aims to saturate a critical neuron in the SNN, causing a drastic accuracy drop \textbf{only when triggered}.
The HT should remain dormant and undetected until the adversary (or a third-party attacker) applies the predefined input trigger.
Once activated, the HT can disrupt the system's performance or jeopardize the client's application at a chosen time.

\section{Proposed Attack}\label{sec:attack}

\subsection{Principle of operation}
\label{sec:attack_principle}

\begin{figure}[t]
    \centering  \includegraphics[width=0.78\columnwidth]{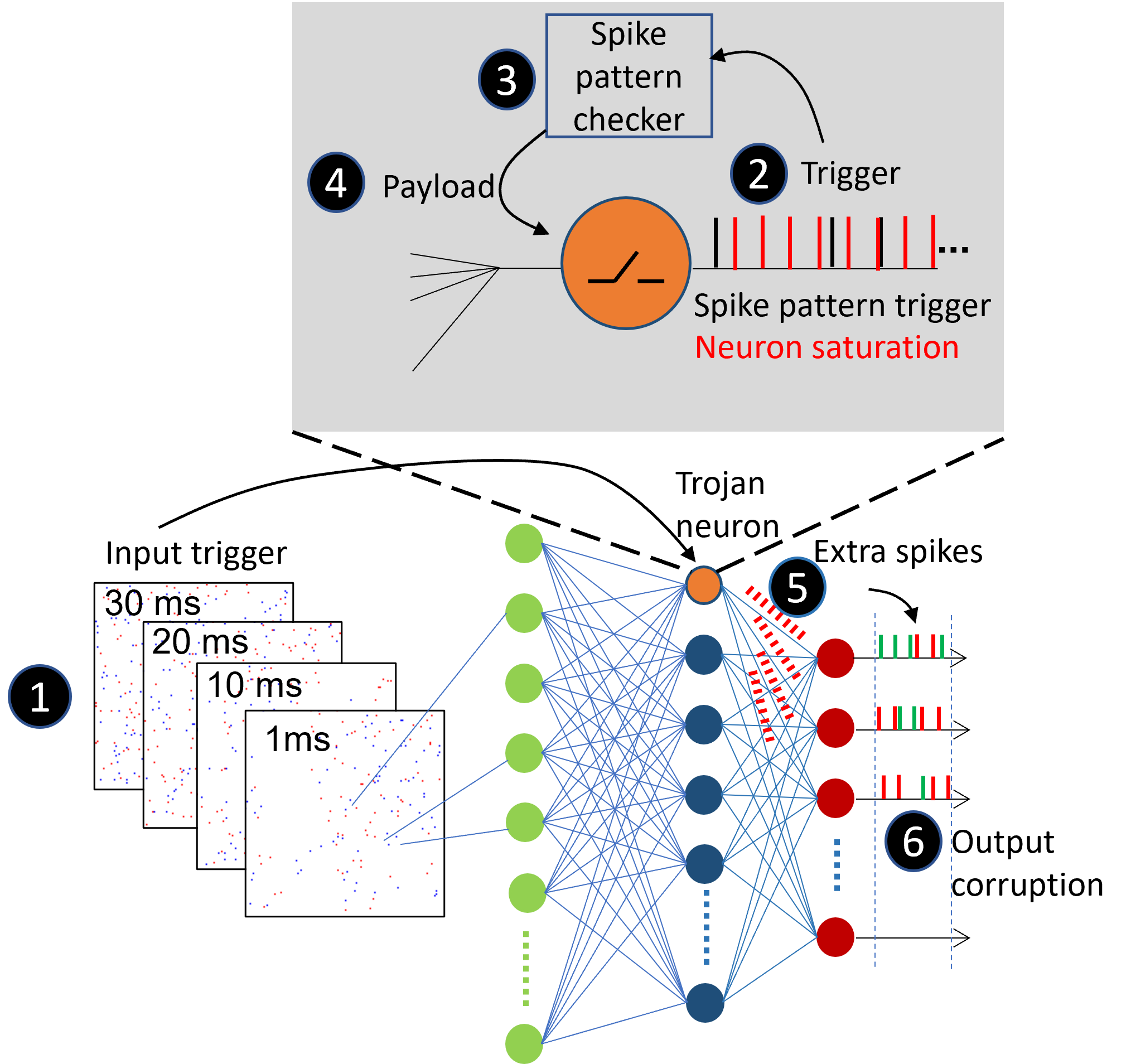}
    \caption{Operating principle of proposed attack.}\vspace{-0.0cm}
    \label{fig:attack}
\end{figure}

We exploit a vulnerability that has its origin in spiking neuron faults and the error behavior they can produce at the network's output \cite{SE-SAC-ML-BS21, TCWL21, PuHaSh22,E-SSCS23,RaSt25}. The attack principle is illustrated in Fig. \ref{fig:attack}. The HT design is condensed in the area of one victim neuron, which we refer to as Trojan neuron. The Trojan neuron is selected such that when it behaves erroneously the accuracy of the network drops drastically. The faults that have the largest impact are dead neuron faults, i.e., the neuron halts processing any incoming spikes having a constant zero-spike output, and saturated neuron faults, i.e., the neuron fires spikes all the time, even without any external stimuli \cite{E-SSAC-ML-BS20}. Herein, we consider a HT payload mechanism that forces the Trojan neuron to saturate, as this type of fault is the most lethal. 
This is because an always-on neuron pollutes the network with non-stop spikes, which can spread out quickly modifying the number or timing of output spikes based on which the network makes a decision. 
This HT payload choice will be justified with simulation experiments in Section \ref{sec:Trojan_neuron_selection_results}.
In Sections \ref{sec:analog_neuron} and \ref{sec:digital_accelerator}, we demonstrate how neuron saturation can be implemented in both analog spiking neurons and digital SNN accelerators with negligible area and power footprint. 

To select the Trojan neuron, the attacker performs a neuron fault injection experiment \cite{GMCRSC24, SpHaSt24} to identify the most critical neuron, as described in Section \ref{sec:fault_simulation}. Note that the output neurons are inherently critical. For example, in a classification application, there is one neuron per class. If a neuron saturates, the corresponding class is always selected.
Given the relatively small number of neurons in the output layer, defenders can easily analyze and identify potential malicious hardware modifications. Consequently, a Trojan neuron is more effectively concealed within the larger hidden layers.

The HT triggering mechanism is a spike pattern checker that monitors the output of the Trojan neuron. The checker looks up for a specific spike pattern trigger and when it occurs the HT payload is applied to the Trojan neuron. This spike pattern is selected such that it is not met in normal operation so as to avoid accidental activation of the HT. This can be checked on the available dataset, but, in general, given the sparsity of spikes, using a spike pattern with many spikes will ensure that it will not be met. The spike pattern checker circuit design is described in Section \ref{sec:spike_pattern_checker}.

The HT is triggered externally. In Section \ref{sec:input_trigger}, we propose an algorithm to craft a spiking input that produces the desired spike pattern trigger at the output of the Trojan neuron. The algorithm uses gradient back-propagation to optimize an input with the loss function defined as the difference between the output of the Trojan neuron and the spike pattern trigger. The input can be viewed as a sequence of frames, with each frame having a number of pixels. A pixel can have a zero value or can carry a spike. Fig. \ref{fig:attack} shows an example input from a vision dataset with blue (red) dots representing excitatory (inhibitory) spikes with positive (negative) polarity that increase (decrease) the membrane potential of the receiving neuron, making it more (less) likely to reach its threshold and spike.

We note the difference between this input trigger with respect to adversarial examples and triggers in backdoored models. In our case, the input trigger generates a spike pattern at the output of a pre-selected Trojan neuron. Adversarial attacks involve crafting small, carefully designed perturbations to an input that fools a trained model into making incorrect predictions. In backdoor attacks, malicious samples containing a trigger, i.e., images with a patch in the case of image recognition, are included in the dataset at training time. After training, a backdoor model correctly performs the task while makes incorrect predictions when the input contains the trigger. Therefore, to generate the proposed input trigger we require only the trained model, while adversarial attacks require in addition the dataset and backdoor attacks poison the dataset during training.

Fig. \ref{fig:attack} summarizes the sequence of events when the attack is launched. The input trigger generates the spike pattern trigger at the output of the Trojan neuron. The spike pattern checker detects this trigger and the payload is applied to the Trojan neuron causing irreversible neuron saturation. The extra non-stop spikes that are generated due to the saturation mode propagate through the subsequent layers polluting the output spike trains and, thereby, leading to incorrect predictions.

Revisiting our threat model, the HT is implanted by the SNN hardware accelerator provider, e.g., the main adversary, during the design stage. At this stage, the adversary randomly selects a neuron to serve as the Trojan neuron and inserts the HT trigger and payload mechanisms. When the victim outsources the SNN model and dataset to the adversary for training, the adversary identifies the critical neuron in the model through a fault injection campaign, selects the spike pattern trigger, and generates the input trigger. The compiler used to map the SNN model onto the hardware is manipulated such that the critical neuron is mapped onto the hardware Trojan neuron. Finally, the selected spike pattern trigger is programmed on chip. The adversary sends the Trojan-infected SNN hardware accelerator to the inconspicuous victim that can be compromised at any time by applying the input trigger. The adversary can disclose the input trigger to a third-party attacker upon request.

\subsection{Critical Neuron Identification} \label{sec:fault_simulation}

To select the Trojan neuron, the attacker can rely on a fault simulation experiment. Neuron saturation is known to be the worst-case, most catastrophic fault type, as discussed in Section \ref{sec:attack_principle}. Thus, the attacker can focus the analysis on this fault type. In Section \ref{sec:Trojan_neuron_selection_results}, we perform fault injection for both saturated and dead neuron faults, confirming our hypothesis that saturated neuron faults are more lethal. The fault simulation experiment consists in cycling over all neurons and for one neuron at a time injecting a fault and performing inference on the complete dataset to assess the impact of the fault on the network's accuracy. 

In our experiment, fault injection is performed in the Spike LAYer Error Reassignment (SLAYER) framework \cite{shor18} by customizing the flow of computations. In SLAYER, the output of a layer is generated for a duration equal to the duration of the input before it is passed to the next layer. A dead neuron fault is emulated by setting the output spikes of the target neuron to zero before passing the layer's output to the next layer. Accordingly, a saturated neuron fault is emulated by forcing the target neuron to fire a spike at every time step of the global clock.

A fault is labeled as critical if one or more samples of the dataset are misclassified while previously they were correctly classified by the fault-free network, otherwise it is labeled as benign. A significant portion of faults end up being benign, whereas each critical fault results in a different accuracy drop. The analysis can be performed per layer to identify the most critical neuron per layer, i.e., the one which if it gets saturated causes the most detrimental effect on the network. These neurons become candidates for serving as the Trojan neuron. As already mentioned in Section \ref{sec:attack_principle}, to evade detection, it is suggested to place the Trojan neuron in one of the dense hidden layers.

\subsection{Input Trigger Generation}\label{sec:input_trigger}

We interpret the input as being decomposed into frames of size $W \times H$ over time and we denote it by $I(t_k,x_{ij})$, where $t_k$ denotes a discrete time-step and $x_{ij}$ denotes the spatial location on the frame, $i=1, \cdots, W$ and $j= 1, \cdots, H$. If the input has duration $T$ and $T_f$ denotes the global clock period, then $k=1,\cdots,T/T_f$. At time-step $t_k$, $I(t_k,x_{ij})=1$ if $x_{ij}$ carries a spike, otherwise $I(t_k,x_{ij})=0$ denotes no spike.

Let $O(t)$ denote the output spike train of the Trojan neuron, which is a binary vector, with $O(t_k)=1$ corresponding to a spike and $O(t_k)=0$ to no spike. 



\subsubsection{Spike pattern trigger selection}

Let the spike pattern trigger be denoted by $P(1:d) \in \{0,1\}^d$, where $d$ denotes its length, with $P(i)=1$ corresponding to a spike and $P(i)=0$ to no spike. Let $[t_{\alpha},t_{\beta}]$ be the time window when the matching between the Trojan neuron output and $P$ is attempted, where $1 \leq \alpha,\beta \leq T/T_f$, $\alpha < \beta$, and $\beta-\alpha+1=d$. In our implementation, without loss of generality, we define this time window to be $[t_{\alpha},t_{\beta}]=[1+T/T_f-d,T/T_f]$, i.e., we consider the last $d$ time-steps of the inference window.

The attacker needs to ensure that the probability of $P$ occurring in normal operation is practically negligible. For this purpose, the complete dataset is applied to the SNN and the output $O^i$ of the Trojan neuron is recorded, where $i$ denotes the input sample index. $P$ must have a Hamming distance of minimum one from every $O^i \left ( \left ( 1+T/T_f-d\right ):T/T_f \right )$:\vspace{-0.1cm}

\begin{equation}\label{eq:HD}
    d_H \left ( O^i \left ( \left ( 1+T/T_f-d \right ):T/T_f \right ) ,P \right ) \geq 1, \forall i. \vspace{-0.1cm}
\end{equation}

Another condition for selecting $P$ is related to the refractory period $\tau_\text{ref}$ of spiking neurons, referring to a period immediately following the generation of a spike during which the neuron is unable to fire another spike. This mechanism helps regulate the firing frequency of the neuron. It imposes that there should be $\tau_{ref}$ \texttt{0}s between two consecutive \texttt{1}s in $P$. This condition can be mathematically represented as:\vspace{-0.2cm}

\begin{equation}\label{eq:refractoriness}
    P(i)*P(j)=0, \forall j \in \{i+1,…,i+\tau_{ref}\}.
\end{equation}

$P$ is selected to be a minimum length pattern that satisfies the conditions in Eqs. (\ref{eq:HD})-(\ref{eq:refractoriness}).

\subsubsection{Input trigger generation algorithm}

The objective of the input trigger generation algorithm is to craft an input $I_{tr}$ such that $d_H \left ( O\left ( \left ( 1+T/T_f-d \right ):T/T_f \right ) , P \right )=0$. We formulate the following optimization problem:\vspace{-0.1cm}

\begin{equation}\label{eq:opt}
   I_{tr}=\arg\min_{I, T} d_H \left ( O\left ( \left ( 1+T/T_f-d \right ):T/T_f \right ), P \right ). \vspace{-0.1cm}
\end{equation}

For the optimization we use the Adam optimizer \cite{Adam2017optimization}. The challenge is to deal with the non-differentiable spike events that prevent the direct use of traditional gradient-based optimization. For this purpose, we use the Gumbel-Softmax \cite{Gumbel_2_2017, Gumbelsoftmax2017} and Straight Through Estimator (STE) \cite{ste2013} techniques.
More specifically, we consider a randomly initialized real-valued input $I_{real}$. It is first passed through the Gumbel-Softmax function that provides a differentiable approximation of a binary input, $I_{soft}$, in the range $(0,1)$:\vspace{-0.1cm}
    
\begin{equation}\label{eq:GumbelSoftmax}
    I_{soft} = GumbelSoftmax(I_{real}, \tau),\vspace{-0.1cm}
\end{equation}
    
\noindent where $\tau$ is the temperature parameter of the Gumbel-Softmax function that controls the sharpness of the approximation. Lower temperature leads to values closer to binary, whereas higher temperature leads to values closer to each other. In our algorithm, this parameter is adaptive. In order to perform the forward pass through the SNN model, we need to convert $I_{soft}$ into the binary input $I$. This is achieved with the STE function that binarizes $I_{soft}$ using a threshold $0.5$:\vspace{-0.1cm}

\begin{equation}\label{eq:STE}
    I = STE(I_{soft}).\vspace{-0.1cm}
\end{equation}

\noindent A forward pass is performed to compute the output $O$ of the Trojan neuron and subsequently the objective function: 

\begin{equation}\label{eq:loss_function}
    \mathcal{L} =d_H \left ( O\left ( \left ( 1+T/T_f-d\right ):T/T_f \right ) , P \right )
\end{equation}

\noindent Thereafter, in the backward pass we compute the gradient of $\mathcal{L}$ with respect to the input $I$. For this purpose, we use the same back-propagation flow as during the training of the SNN model, i.e., starting from the layer of the Trojan neuron, gradients with respect to the input of the layers are propagated layer by layer moving backward to the input. When we reach
the input, the STE function passes on the incoming gradient as if it was an identity function. In the next step, the Adam optimizer makes corrections to $I_{real}$ using the learning rule:\vspace{-0.1cm}

\begin{equation}\label{eq:learning}
    I_{real} \leftarrow I_{real} - lr*\nabla_{I_{real}} \mathcal{L},\vspace{-0.1cm}
\end{equation}

\noindent where $lr$ is the learning rate.

\begin{algorithm}[t]
\scriptsize
\caption{Input trigger generation pseudo-algorithm}\label{alg:pseudocode}

\KwData{SNN model, dataset, global clock period $T_f$, Trojan neuron, neurons' refractory period $\tau_\text{ref}$, maximum optimization time $t_{limit}$}
\KwResult{Input trigger $I_{tr}$}

$\text{Perform full inference and record Trojan neuron outputs $O^i$}$\;

$\text{Define $P$ of minimum length $d$ that satisfies the conditions in Eqs. (\ref{eq:HD})-(\ref{eq:refractoriness})}$\;

$\text{Randomize a real-valued input $I_{real}$ with duration $T=d \times T_f$}$\;
$t \gets \text{current time}$\;
$t_{limit} \gets \text{current time} + t_{limit}$\;

\While{$\mathcal{L} \neq 0$ $\land$ $t < t_{limit}$}{

$\textrm{Generate the binary input $I$ from $I_{real}$ using Eqs. (\ref{eq:GumbelSoftmax})-(\ref{eq:STE})}$\;

$\textrm{Perform a forward pass and calculate $\mathcal{L}$ in Eq. (\ref{eq:loss_function})}$\;

\If{$\mathcal{L}=0$}{
$I_{tr} \gets I$\;
}

$\textrm{Perform a backward pass and use Eq. (\ref{eq:learning}) to refine $I_{real}$}$\;

$t \gets \text{current time}$\;
}

$\textrm{If $\mathcal{L} \neq 0$, then repeat with $T=T+1$ and/or use a more sparse $P$}$\;

\end{algorithm}

Algorithm \ref{alg:pseudocode} shows the complete input trigger generation pseudo-algorithm. If there is no improvement of the objective function after some time $t_{limit}$, to help convergence, we repeat the algorithm by increasing the input duration by one time step and/or making the trigger pattern sparser while still satisfying the Hamming distance and refractory period conditions.

\section{Simulation results}\label{sec:results}

\subsection{Case studies}\label{sec:case_studies}

\begin{figure}[t]
\centering
\includegraphics[width=0.9\linewidth]{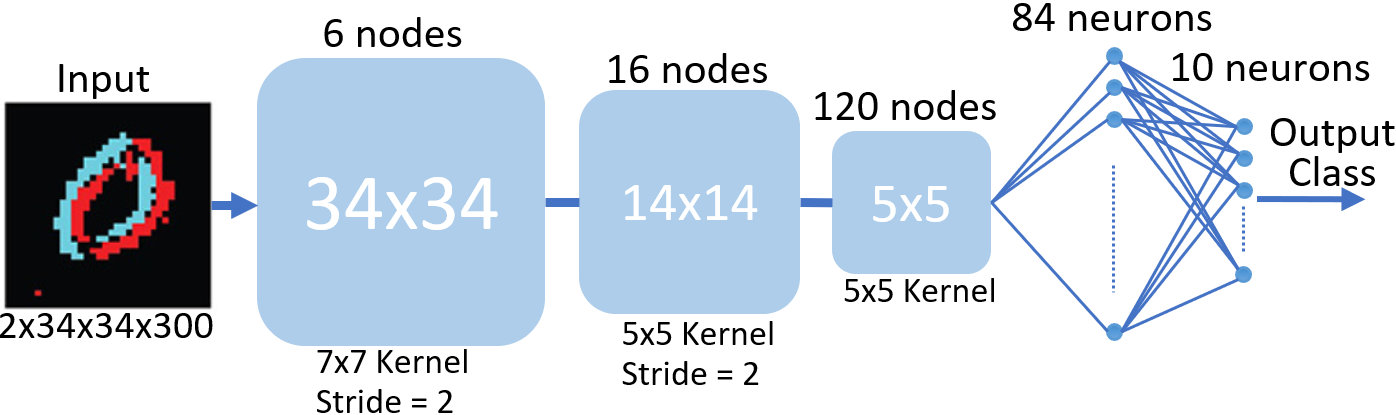}
\caption{SNN architecture for the NMNIST dataset.}\vspace{-0.2cm}
\label{fig:nmnist_SNN}
\end{figure}

\begin{figure}[t]
\centering
\includegraphics[width=1.0\linewidth]{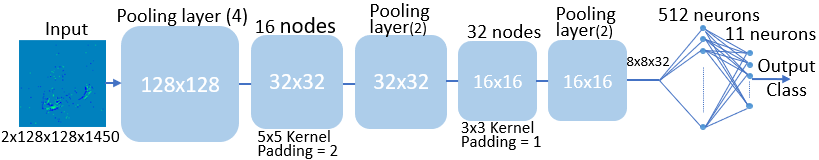}
\caption{SNN architecture for the IBM DVS128 Gesture dataset.}\vspace{-0.2cm}
\label{fig:IBM_SNN}
\end{figure}

\begin{figure}[t!]
\centering
\includegraphics[width=0.75\linewidth]{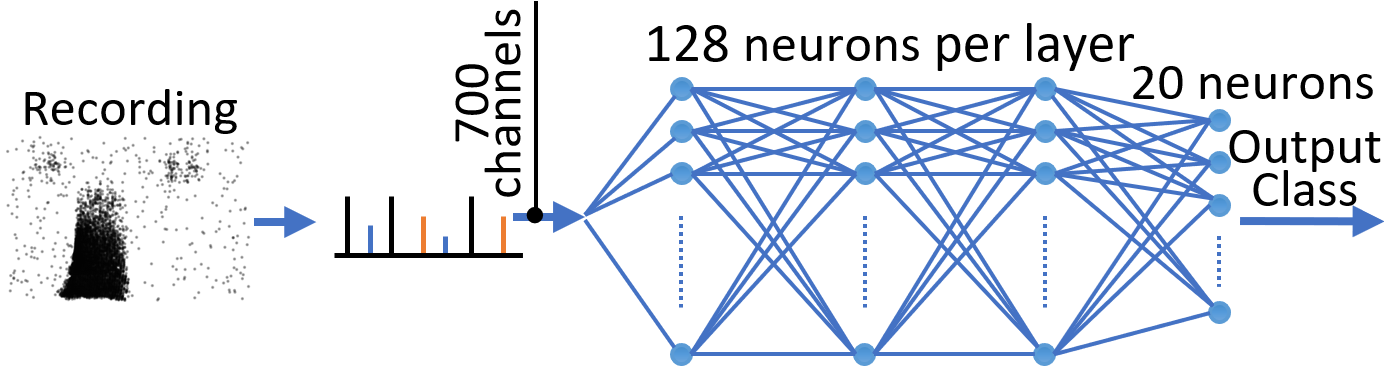}
\caption{SNN architecture for the SHD dataset.}\vspace{-0.0cm}
\label{fig:SHD_SNN}
\end{figure}

The attack is demonstrated by simulation at the application level using three common benchmark datasets in the neuromorphic domain, namely the NMNIST \cite{OJCT15}, IBM DVS128 Gesture \cite{ATBM17}, and Spiking Heidelberg Digits (SHD) \cite{heidelberg} datasets. 
NMNIST is a spiking version of the original frame-based MNIST dataset containing $70K$ images of handwritten digits from 0 to 9. It was produced by moving a Dynamic Vision Sensor (DVS) while it views MNIST images on an LCD monitor. The IBM DVS128 Gesture dataset was produced by a DVS capturing 11 different hand and arm gestures performed by 29 individuals under 3 different lighting conditions. The SHD dataset consists of 10420 audio recordings of spoken digits from 0 to 9 in German and English languages converted into spike trains. The SNN model architectures for the classification of the three datasets are shown in Figs. \ref{fig:nmnist_SNN}, \ref{fig:IBM_SNN} and \ref{fig:SHD_SNN}. The SNNs for the MNIST and IBM DVS128 Gesture datasets are convolutional, while the SNN for the SHD dataset is fully-connected. The training of the SNN models was conducted using the SLAYER framework \cite{shor18}. Training and inference are accelerated on an NVIDIA A100 GPU. Table \ref{tab:SNN_characteristics} summarizes the main SNN characteristics.

\begin{table}[t!]
\centering
\scriptsize
\caption{SNN characteristics.}
\begin{tabular}{ | m{2.6cm} | m{1.2cm}| m{1.5cm} | m{1.2cm} | } 
  \hline
   & \textbf{NMNIST} & \textbf{IBM} & \textbf{SHD}\\
  \hline
  \hline
  Prediction accuracy & $98.19\%$ & $86.36\%$ & $76.59\%$\\  
  \hline
  \# Output classes & $10$ & $11$ & $20$\\ 
  \hline
  \# Neurons & $1790$ & $25099$ & $404$\\ 
  \hline
  Input spatial dimension & $2 \times 34 \times 34$ & $2\times128\times128$ & $700 \times 1 \times 1$\\ 
  \hline
  Input temporal dimension & $300 \, ms$ & $1.45 \, s$ & $~1 \, s$\\ 
  \hline
  Size training set & $60K$ & $1080$ & $8332$\\ 
  \hline
  Size testing set & $10K$ & $261$ & $2088$ \\ 
  \hline
\end{tabular}\vspace{-0.2cm}
\label{tab:SNN_characteristics}
\end{table}

\subsection{Trojan neuron selection}\label{sec:Trojan_neuron_selection_results}

\begin{table}[t]
\centering
\caption{Fraction of critical and benign faults for each fault type.}
\scriptsize
\begin{tabular}{ | m{3.2cm} | m{1.1cm}| m{1.1cm} | m{1.1cm} | } 
  \hline
   & \textbf{NMNIST} & \textbf{IBM} & \textbf{SHD}\\
  \hline
  \hline
  Critical Dead Neuron Faults & $1324$ & $2501$ & $390$\\ 
  \hline
  Critical Saturated Neuron Faults & $1598$ & $22877$ & $404$\\ 
  \hline
  \hline
  Benign Dead Neuron Faults & $466$ & $22598$ & $14$\\ 
  \hline
  Benign Saturated Neuron Faults & $192$ & $2222$ & $0$\\ 
  \hline
  \hline
  Fault Simulation Time & $\sim$ 1 day & $\sim$ 2 days & $\sim$ 0.5 days \\
  \hline
\end{tabular}\vspace{-0.0cm}
\label{tab:fault_simulation_results}
\end{table}

\begin{figure}[h!]
\begin{subfigure}{0.24\textwidth}
  \centering
  \includegraphics[width=0.99\linewidth]{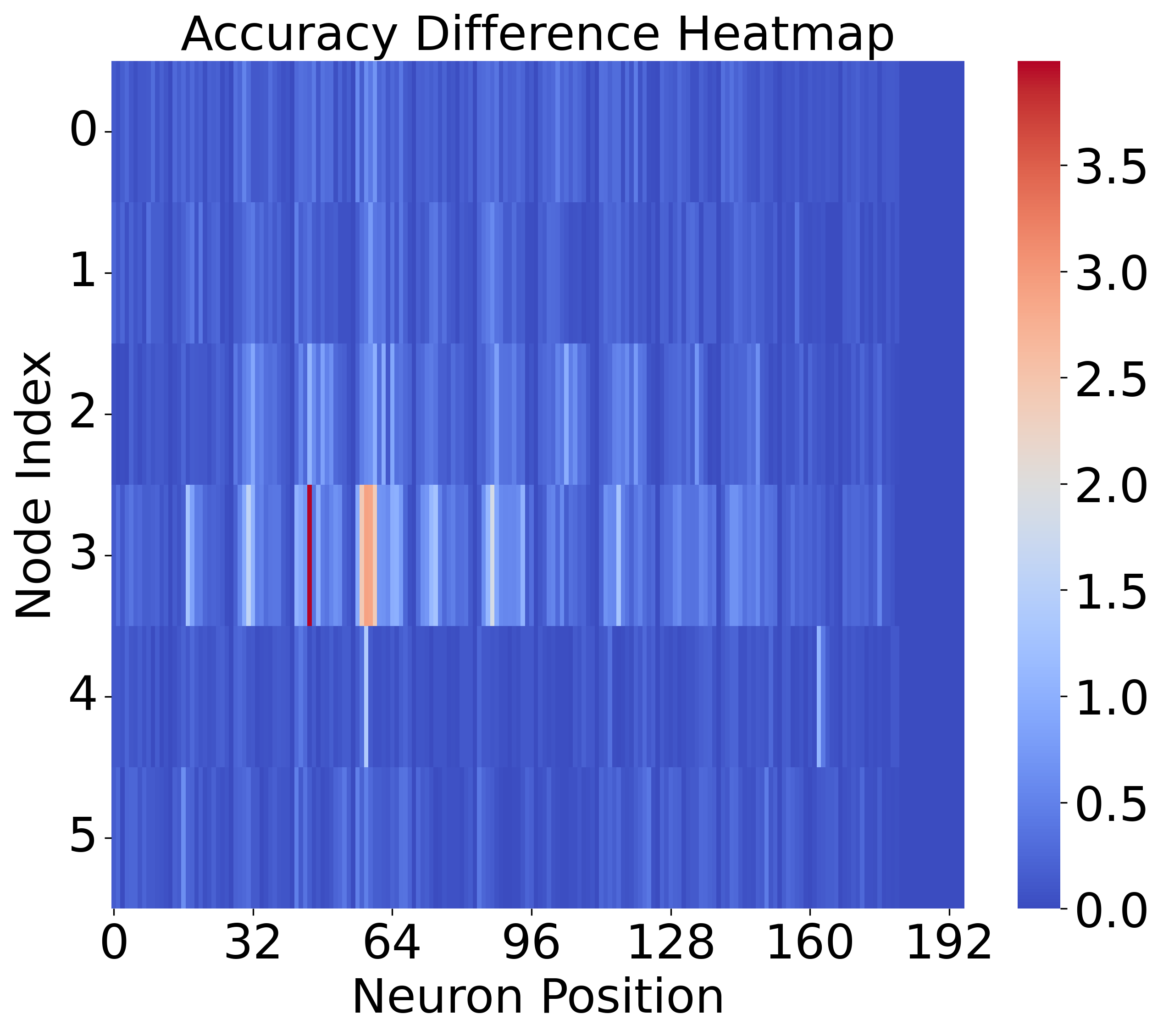}
  \caption{Layer 1-Saturated}
  \label{fig:sfig1}
\end{subfigure}%
\begin{subfigure}{.24\textwidth}
  \centering
  \includegraphics[width=.99\linewidth]{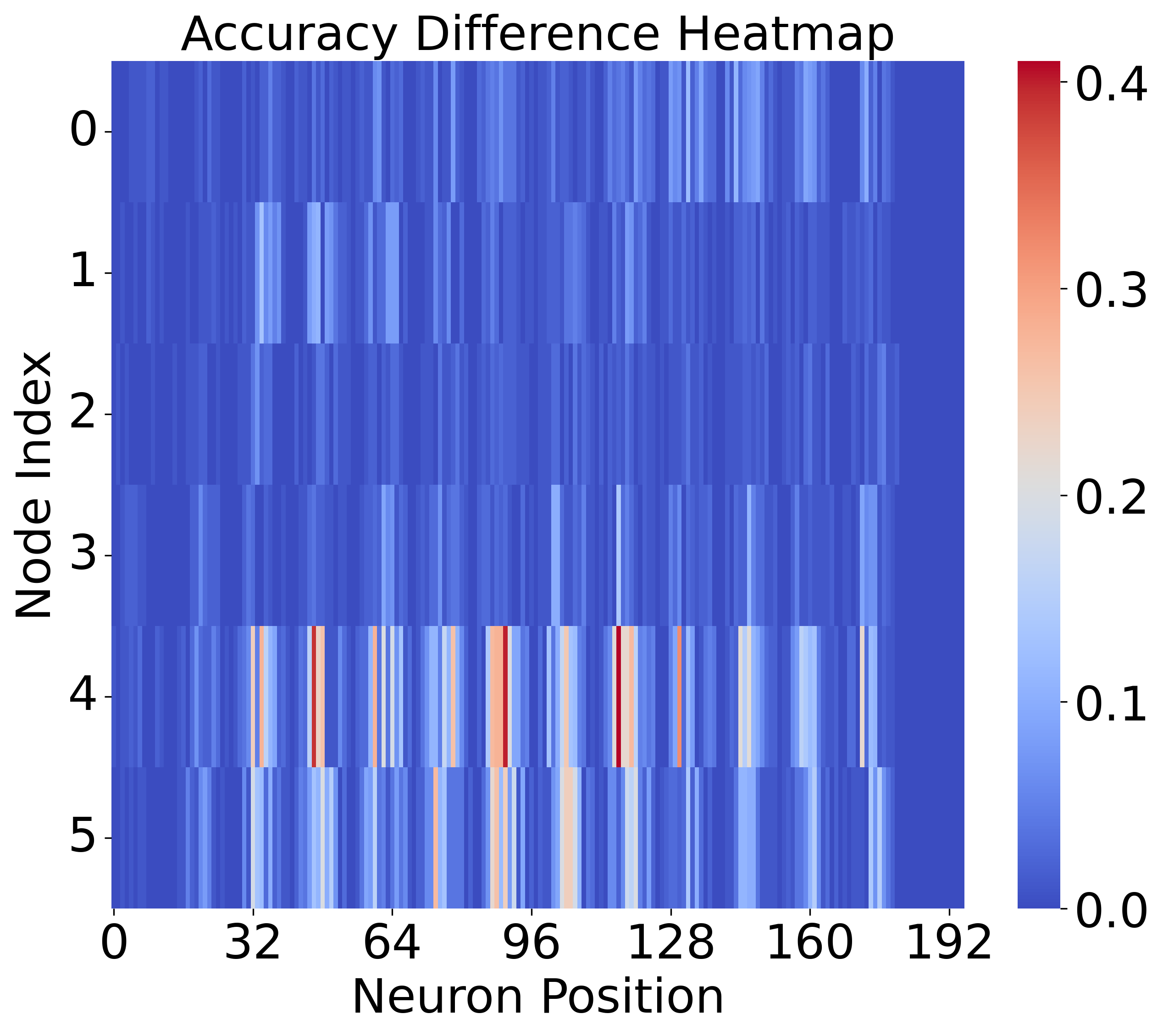}
  \caption{Layer 1-Dead}
  \label{fig:sfig2}
\end{subfigure}
\begin{subfigure}{0.24\textwidth}
  \centering
  \includegraphics[width=0.99\linewidth]{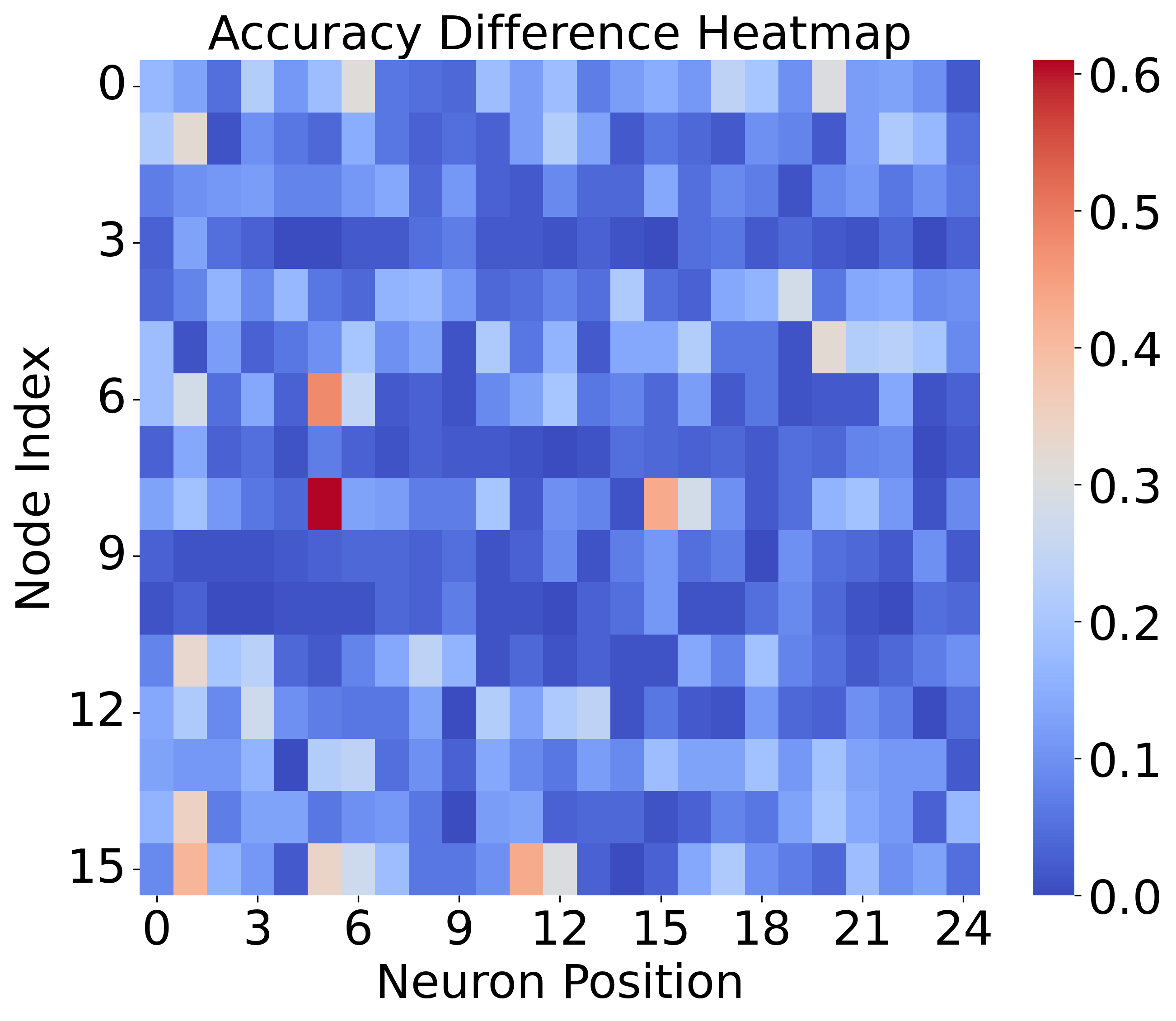}
  \caption{Layer 2-Saturated}
  \label{fig:sfig3}
\end{subfigure}%
\begin{subfigure}{.24\textwidth}
  \centering
  \includegraphics[width=.99\linewidth]{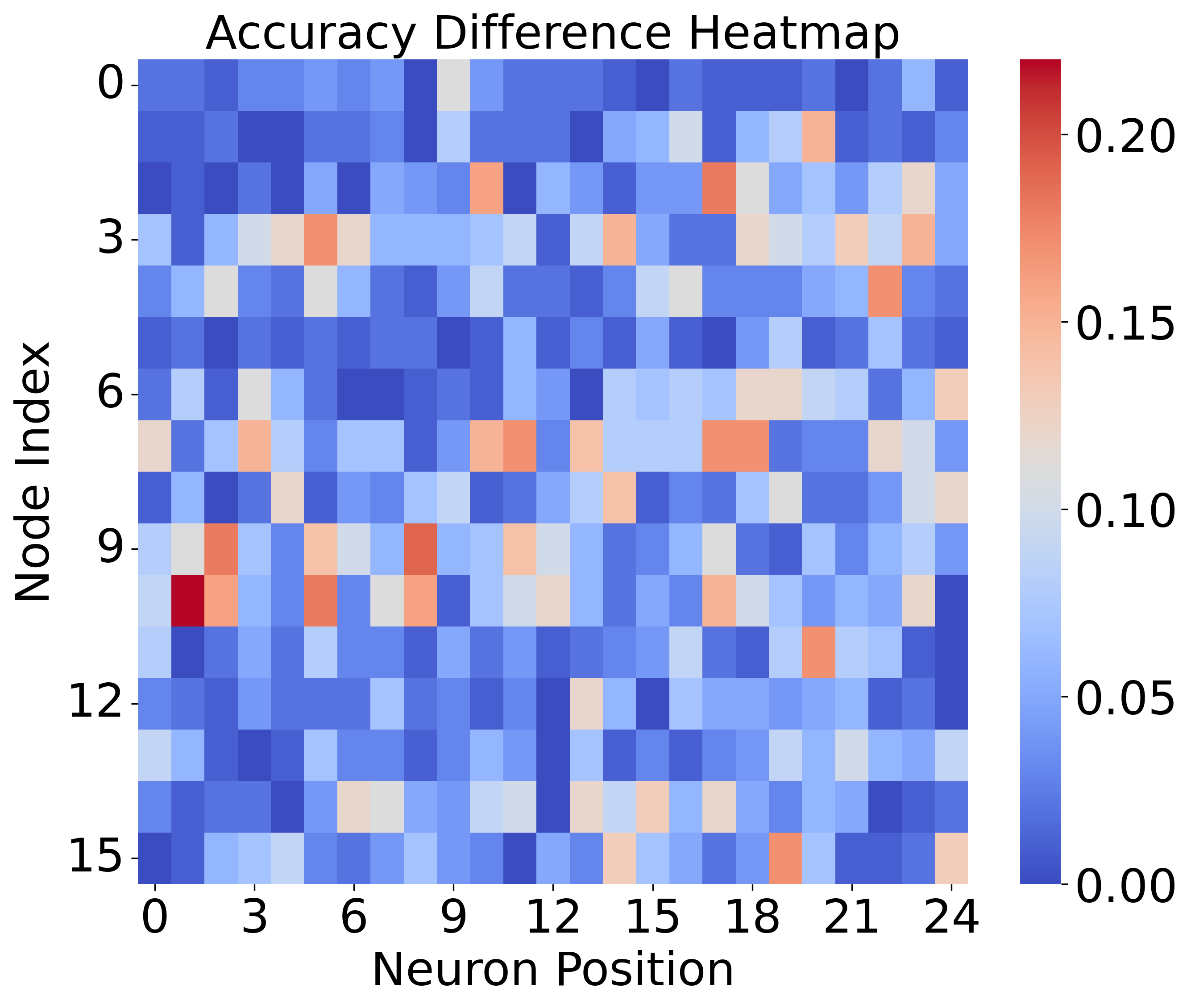}
  \caption{Layer 2-Dead}
  \label{fig:sfig4}
\end{subfigure}
\begin{subfigure}{.24\textwidth}
  \centering
  \includegraphics[width=.99\linewidth]{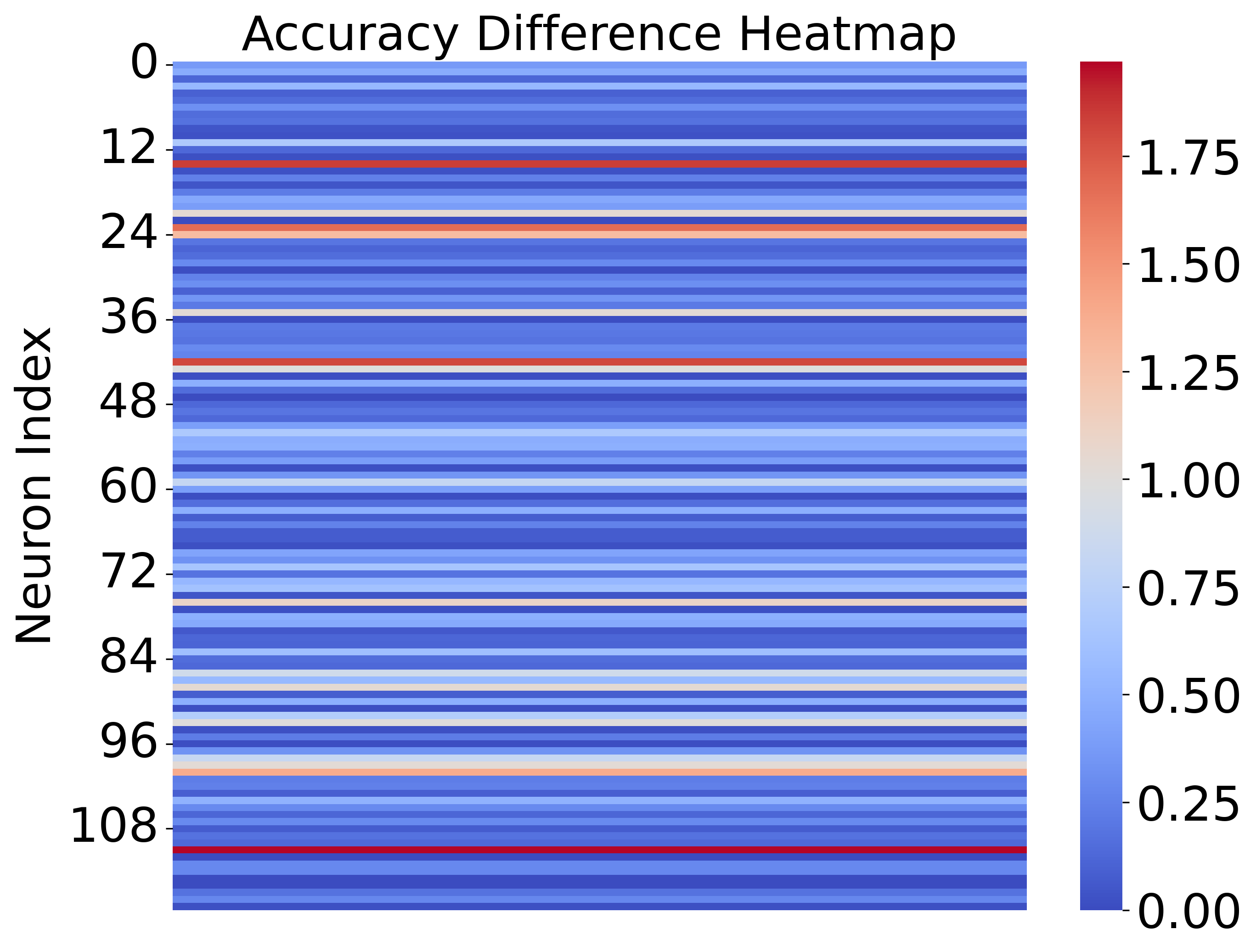}
  \caption{Layer 3-Saturated}
  \label{fig:sfig5}
\end{subfigure}
\begin{subfigure}{.24\textwidth}
  \centering
  \includegraphics[width=.99\linewidth]{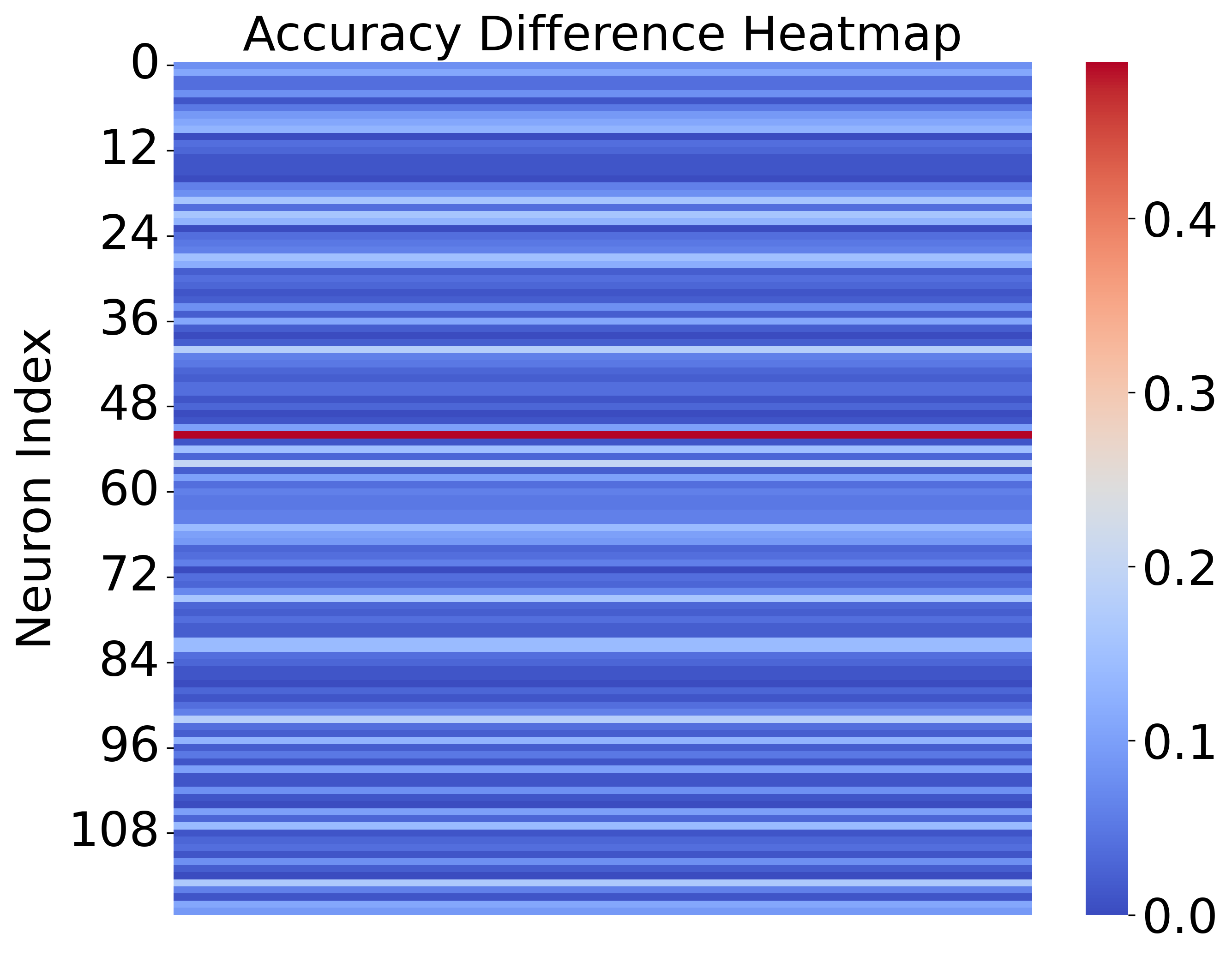}
  \caption{Layer 3-Dead}
  \label{fig:sfig5}
\end{subfigure}
\begin{subfigure}{.24\textwidth}
  \centering
  \includegraphics[width=.99\linewidth]{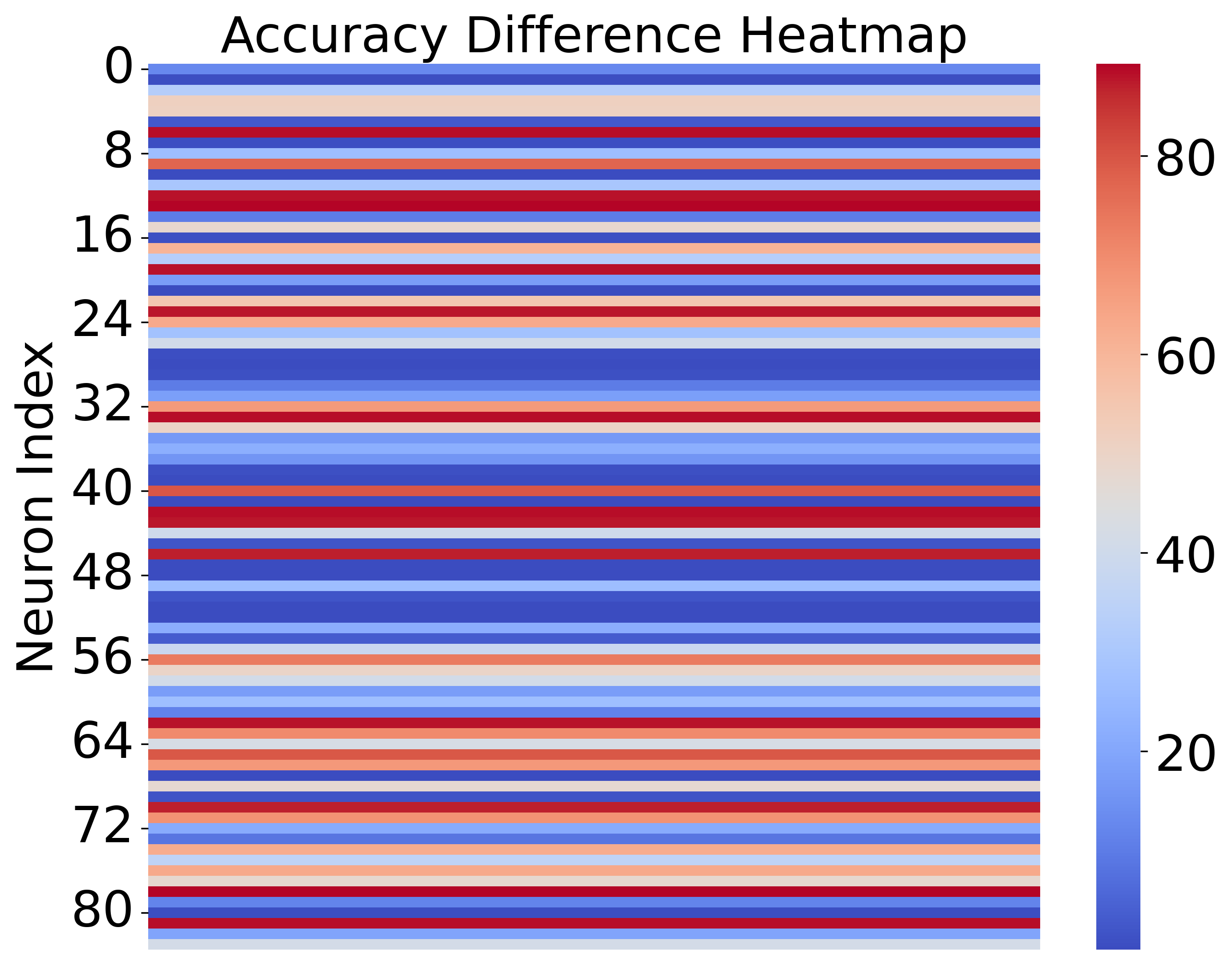}
  \caption{Layer 4-Saturated}
  \label{fig:sfig6}
\end{subfigure}
\begin{subfigure}{.24\textwidth}
  \centering
  \includegraphics[width=.99\linewidth]{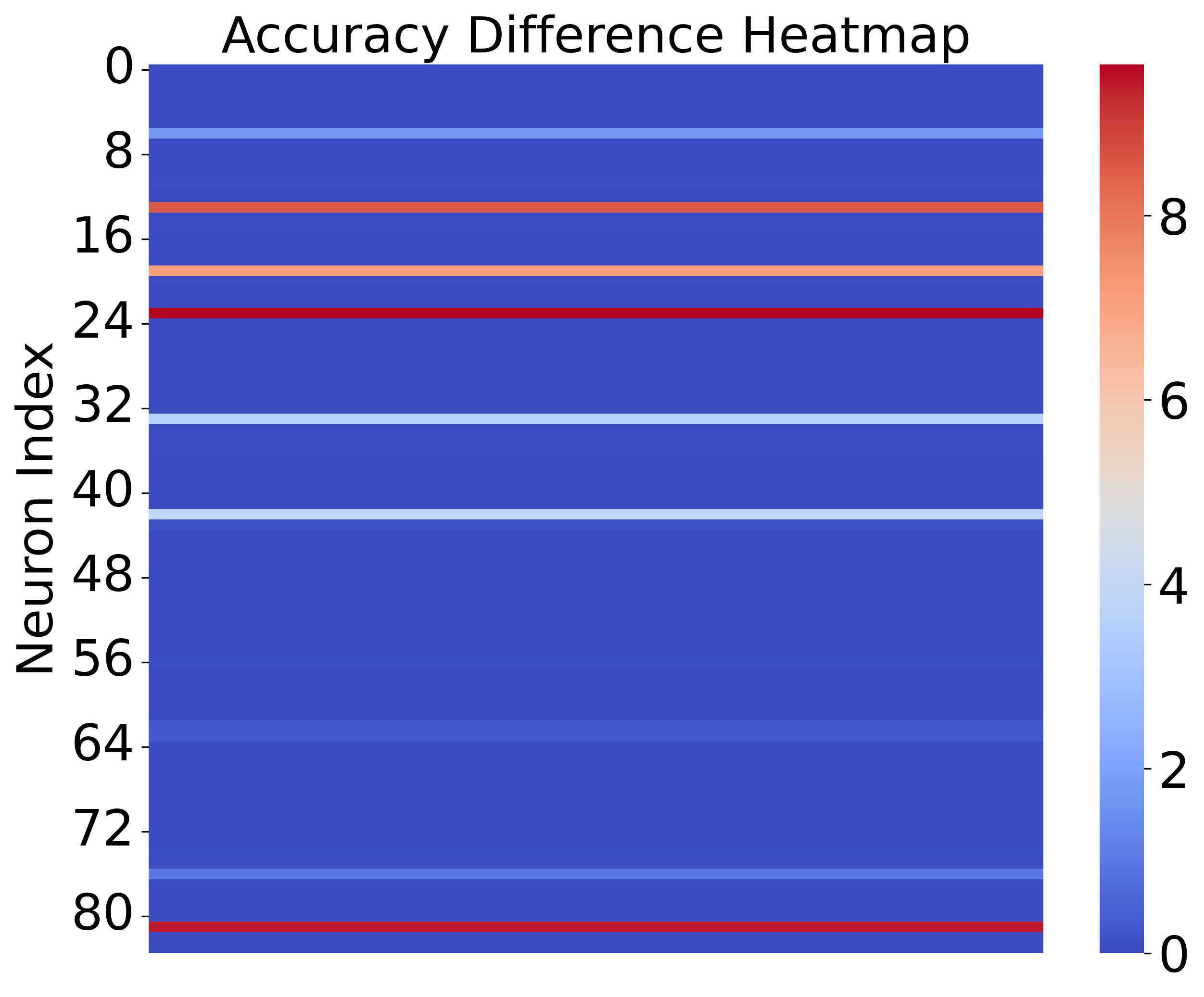}
  \caption{Layer 4-Dead}
  \label{fig:sfig7}
\end{subfigure}
\begin{subfigure}{.24\textwidth}
  \centering
  \includegraphics[width=.99\linewidth]{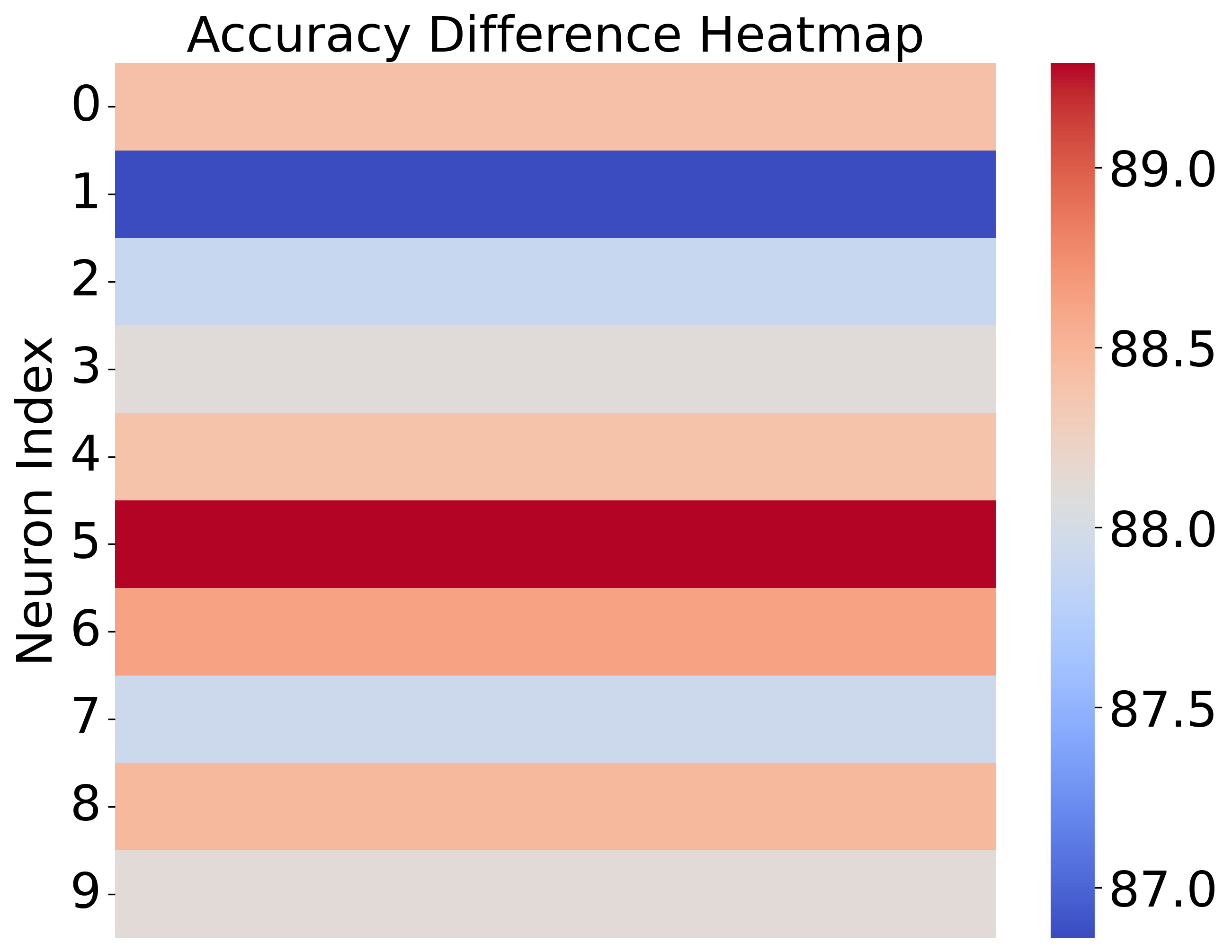}
  \caption{Layer 5-Saturated}
  \label{fig:sfig8}
\end{subfigure}
\begin{subfigure}{.24\textwidth}
  \centering
  \includegraphics[width=.99\linewidth]{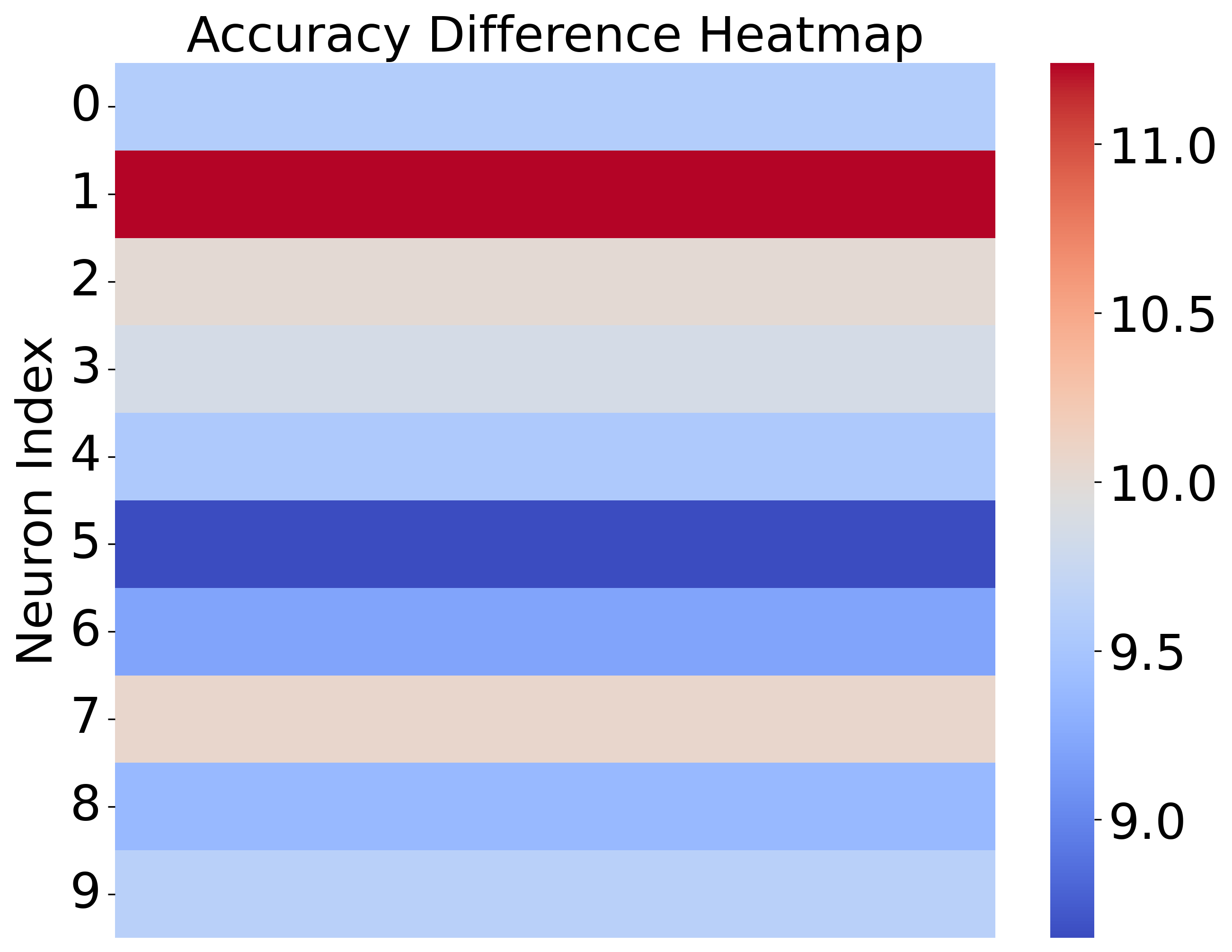}
  \caption{Layer 5-Dead}
  \label{fig:sfig9}
\end{subfigure}
\caption{Accuracy drop per neuron per layer for saturated and dead neuron faults.}\vspace{-0.4cm}
\label{fig:nmnist_faults_2D}
\end{figure}

\begin{table}[t]
\centering
\caption{Maximum accuracy drop noticed across all neurons and fault types for each layer. In all cases, the most critical fault is a saturated neuron fault.}
\scriptsize
\begin{tabular}{ | m{1.cm} | >{\centering\arraybackslash}m{1.3cm}| >{\centering\arraybackslash}m{1.3cm} | >{\centering\arraybackslash}m{1.3cm} | } 
  \hline
   & \textbf{NMNIST} & \textbf{IBM} & \textbf{SHD}\\
  \hline
  \hline
  Layer 1 & $3.99\%$ & $69.32\%$ & $71.42\%$\\ 
  \hline
  Layer 2 & $0.61\%$ & $2.65\%$ & $66.43\%$\\ 
  \hline
  Layer 3 & $1.97\%$ & $77.27\%$ & $70.23\%$\\ 
  \hline
  Layer 4 & $89.29\%$ & $77.27\%$ & $72.08\%$\\ 
  \hline
  Layer 5 & $89.29\%$ & - & -  \\ 
  \hline
\end{tabular}
\label{tab:most_critical_fault_positions}\vspace{-0.4cm}
\end{table}

Table \ref{tab:fault_simulation_results} shows the partition of dead and saturated neuron faults into critical and benign, as well as the total fault simulation time. We observe that the critical saturated neuron faults outnumber the critical dead neuron faults for every SNN. This implies that a neuron becoming dead may not affect the classification accuracy, but if the same neuron becomes saturated then the accuracy may drop. We also observe that, for a given fault type, the critical faults outnumber the benign faults, except for the IBM SNN and dead neuron fault type.

Fig. \ref{fig:nmnist_faults_2D} shows the effect of each fault type on each neuron per layer for the NMNIST SNN. Each sub-plot of Fig. \ref{fig:nmnist_faults_2D} corresponds to one layer. The first two layers are convolutional, whereas the last three are fully-connected. For convolutional layers, the y-axis corresponds to the node (i.e., feature map) index and the x-axis corresponds to the neuron position within the node. For illustration purposes, the node neuron matrix is vectorized, i.e., the neuron positions within a node are flattened into a row. For the fully-connected layers, each row of the plot corresponds to one neuron. In summary, each rectangle in Fig. \ref{fig:nmnist_faults_2D} corresponds to one neuron and the color of the rectangle indicates the classification accuracy drop if the neuron suffers from the fault according to the color map on the right-hand side.

For every SNN, we recorded the most critical fault type at each layer resulting in the maximum accuracy drop. In all cases, the most critical fault type was neuron saturation confirming our choice to use neuron saturation as HT payload. The maximum accuracy drop noticed per layer is summarized in Table \ref{tab:most_critical_fault_positions}. The impact severity of the fault is shown to be higher in the last layers. 

Based on these results, the attacker can choose the Trojan neuron in one of the hidden layers. Looking at Table \ref{tab:most_critical_fault_positions}, for the MNIST SNN, the Trojan neuron can be selected as the most critical neuron in the fourth layer that has 84 neurons, resulting in an accuracy drop of $89.29\%$ when the HT is activated. For the IBM SNN, we can choose the most critical neuron in the first or third layer, resulting in accuracy drop of $69.32\%$ and $77.27\%$, respectively. For the SHD SNN, the Trojan neuron can be placed in any hidden layer as they all have the same number of neurons, and the accuracy drop will be at least $66.43\%$.

\subsection{Input-trigger generation and payload}

\begin{figure}[h!]
\begin{subfigure}{0.25\textwidth}
  \centering
  \includegraphics[width=0.7\linewidth]{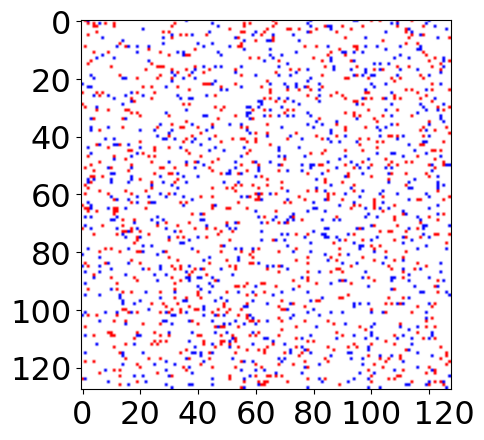}
  \caption{1ms}
\end{subfigure}%
\begin{subfigure}{.25\textwidth}
  \centering
  \includegraphics[width=0.7\linewidth]{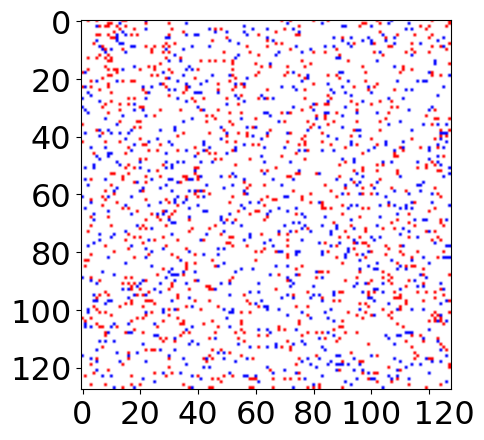}
  \caption{3ms}
\end{subfigure}
\begin{subfigure}{0.25\textwidth}
  \centering
  \includegraphics[width=0.7\linewidth]{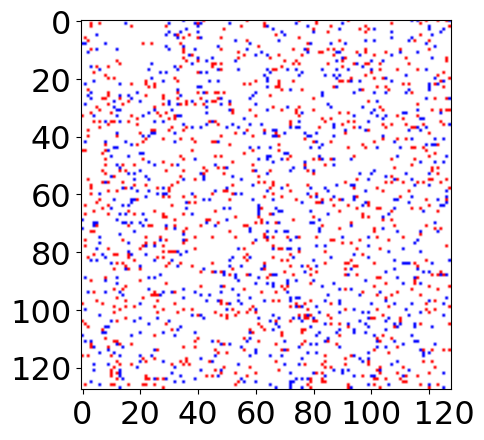}
  \caption{8ms}
\end{subfigure}%
\begin{subfigure}{.25\textwidth}
  \centering
  \includegraphics[width=0.7\linewidth]{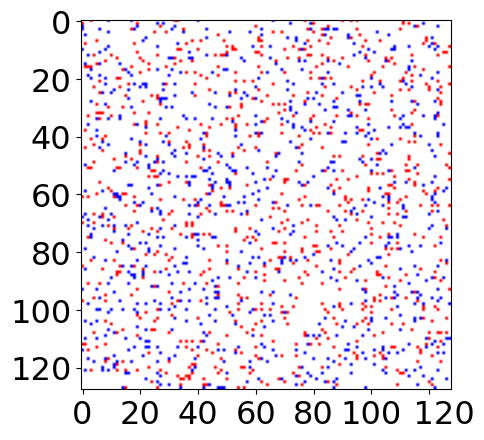}
  \caption{10ms}
\end{subfigure}
\caption{Snapshots of the input trigger.}\vspace{-0.2cm}
\label{fig:ibm_snapshots}
\end{figure}

Without loss of generality, we demonstrate the input trigger generation algorithm for the IBM SNN. We selected the most critical neuron of layer 1 to be the Trojan neuron. Layer 1 is convolutional having 16 nodes of dimension $32 \times 32$ (see Fig. \ref{fig:IBM_SNN}). The specific spatial location of the Trojan neuron is (2,4) in the sixth node. The global clock period is $T_f$=1 ms and the selected trigger pattern at the output of the Trojan neuron is ``1010101010" having duration 10 ms, i.e., a length of $d$=10. The resultant optimized input trigger has duration $T=$15 ms and is composed of 15 frames. The optimization algorithm took around 2 minutes to converge. Fig. \ref{fig:ibm_snapshots} shows four snapshots of the input trigger at 1, 3, 8, and 10 ms, with blue (red) dots indicating spikes with positive (negative) polarity. It was verified that this spike pattern trigger never occurs when passing the 1080+261=1341 available samples in the dataset that have duration 1.45 s (see Table \ref{tab:SNN_characteristics}).

\begin{figure}[t]
\begin{subfigure}{0.25\textwidth}
  \centering
  \includegraphics[width=0.99\linewidth]{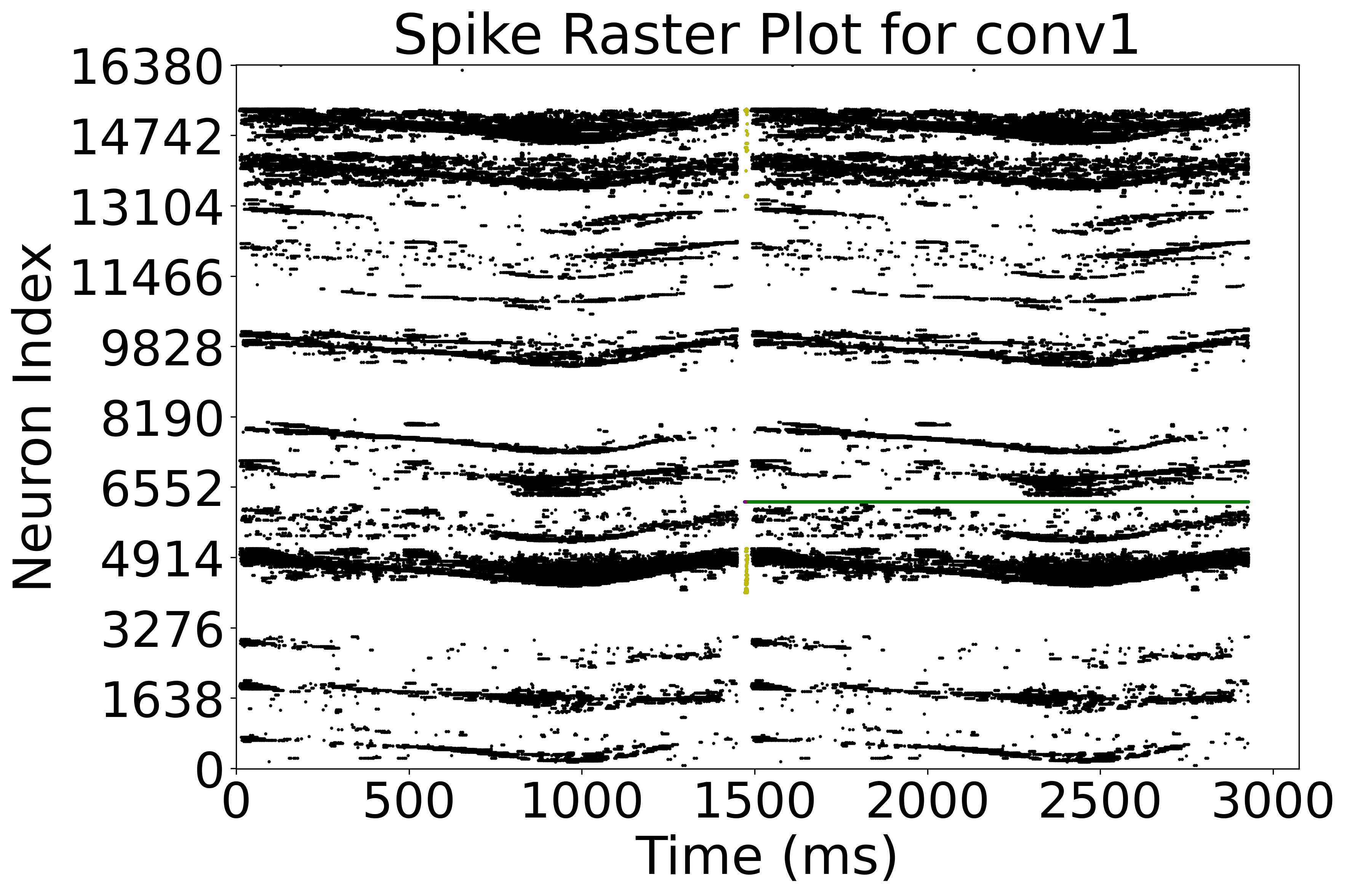}
  \caption{Layer 1}
  \label{fig:sfig1}
\end{subfigure}%
\begin{subfigure}{.25\textwidth}
  \centering
  \includegraphics[width=.99\linewidth]{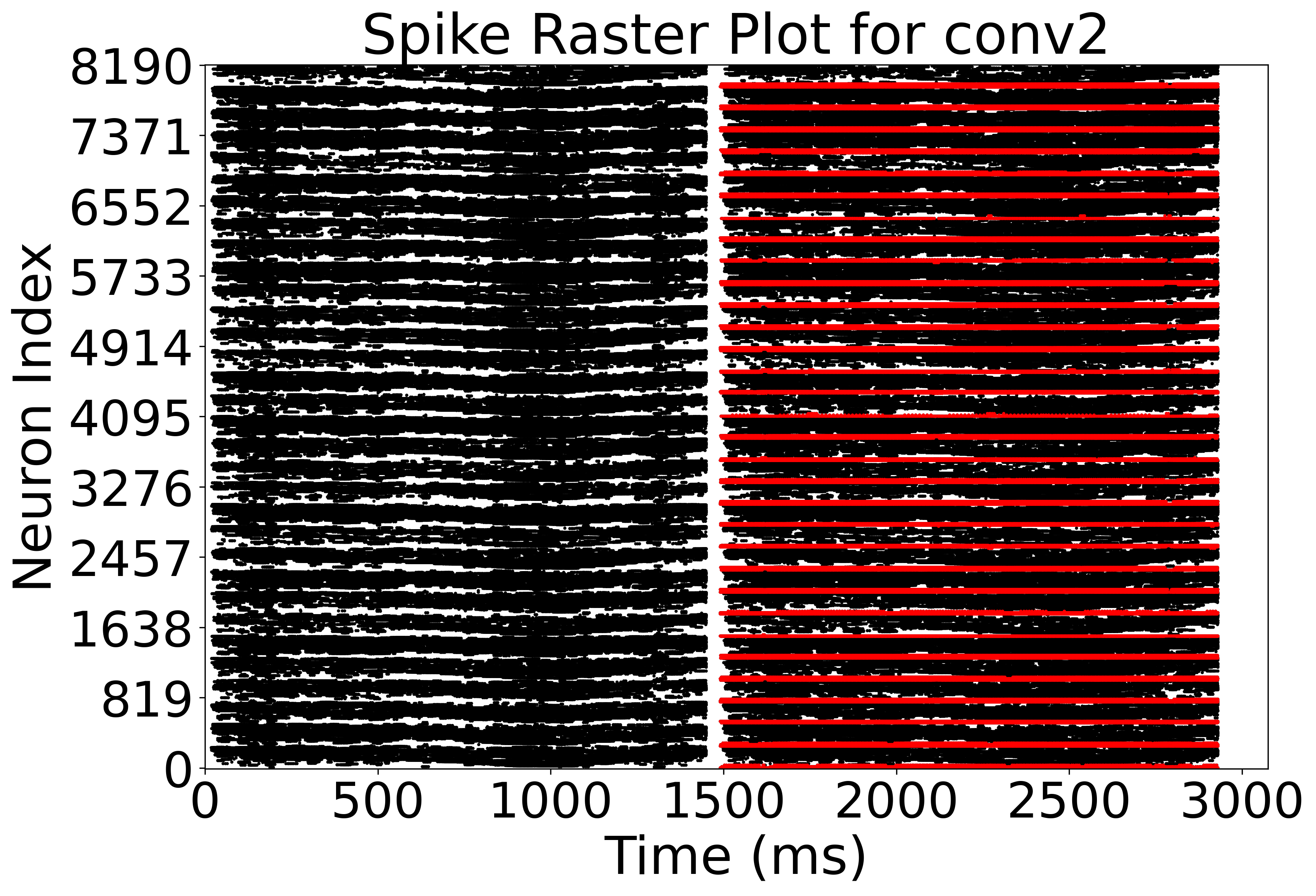}
  \caption{Layer 2}
  \label{fig:sfig2}
\end{subfigure}
\begin{subfigure}{0.25\textwidth}
  \centering
  \includegraphics[width=0.99\linewidth]{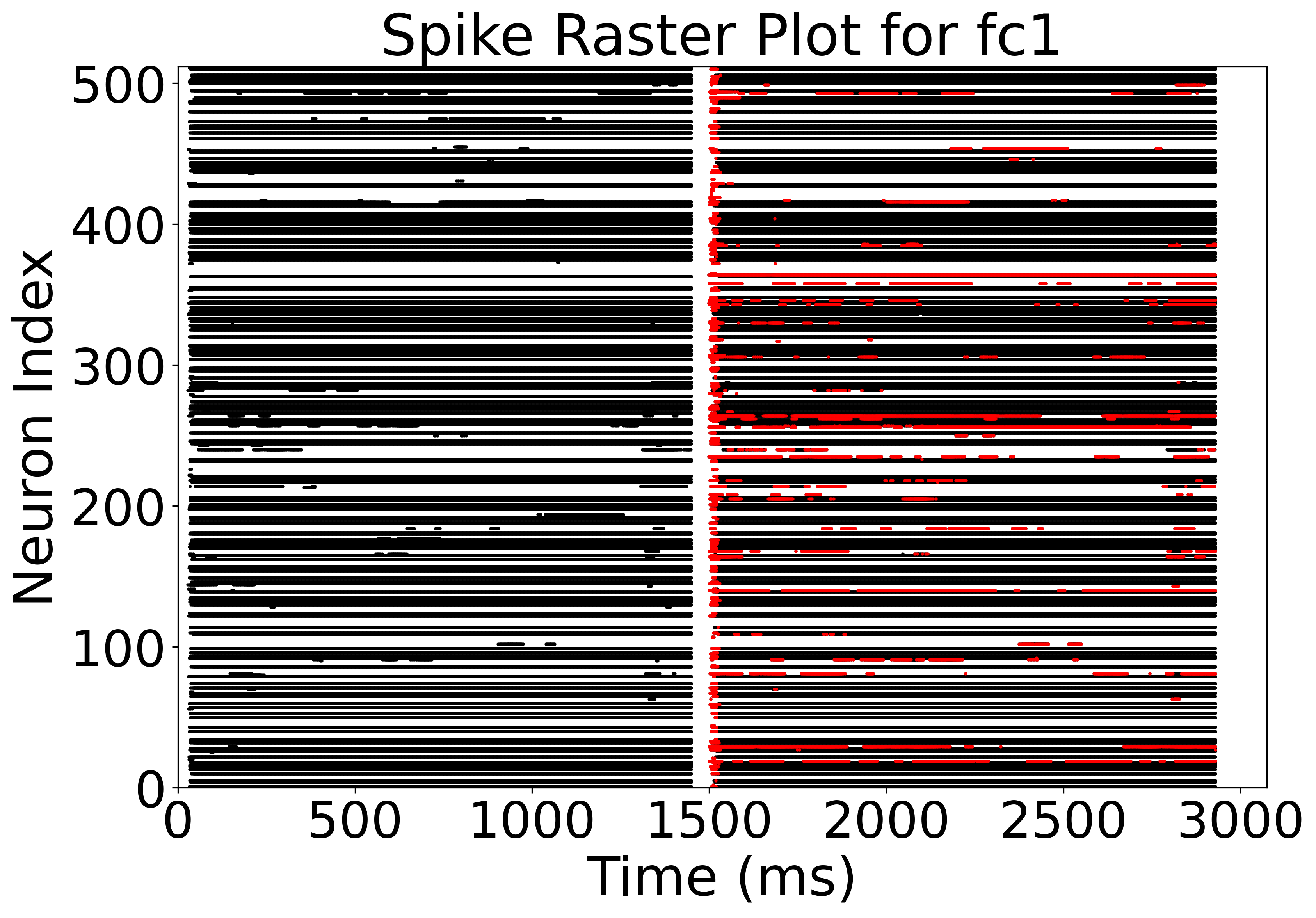}
  \caption{Layer 3}
  \label{fig:sfig3}
\end{subfigure}%
\begin{subfigure}{.25\textwidth}
  \centering
  \includegraphics[width=.99\linewidth]{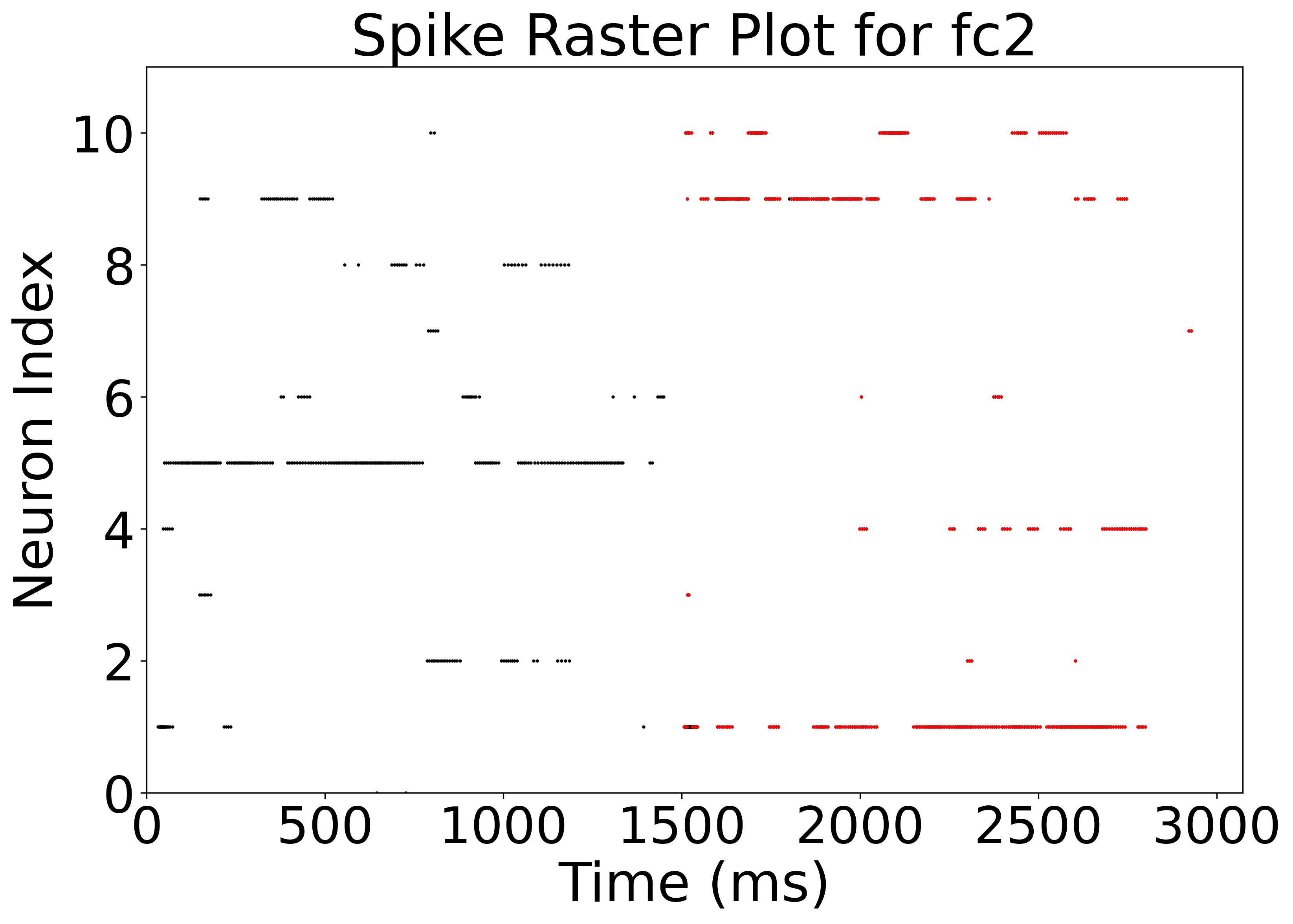}
  \caption{Layer 4}
  \label{fig:sfig4}
\end{subfigure}
\caption{Spiking activity before and after the HT activation for the same dataset sample.}\vspace{-0.4cm}
\label{fig:ibm_trojan}
\end{figure}

As an illustration, we select an input sample that gets misclassified after HT activation. Fig. \ref{fig:ibm_trojan} shows the spiking activity per layer when applying this input sample before and after HT activation. The sequence of applied inputs is as follows: first, the input sample is applied; next, a zero input is introduced to reset the membrane potential of all neurons; then, the input trigger is applied, setting the Trojan neuron to permanent saturation; following this, a sleep mode is repeated; finally, the same input sample is reapplied to observe the effect of the HT. Each plot corresponds to one layer and each row to one neuron. Each row is a raster plot displaying with dots the timing of spike events occurrence for this neuron. The normal HT-free spiking activity is shown on the left-hand side of the plots, the spiking activity when applying the short-duration input trigger is shown in the middle, and the spiking activity after HT activation is shown on the right-hand side. Note that the spiking activity when applying the input trigger is not important as the goal is only to deliver the spike pattern trigger to the Trojan neuron. In layer 1, green dots indicate the Trojan neuron saturation. For all other layers, red dots indicate extra spikes generated due to HT activation. As it can be seen, the HT activation pollutes the network with several extra spikes altering the spike distribution at the output layer 4. The decision is class 5 (left arm clockwise gesture), but after HT activation the winning neuron is the one that corresponds to class 1 (right hand wave gesture).

\begin{figure}[t]
\begin{subfigure}{0.25\textwidth}
  \centering
  \includegraphics[width=0.95\linewidth]{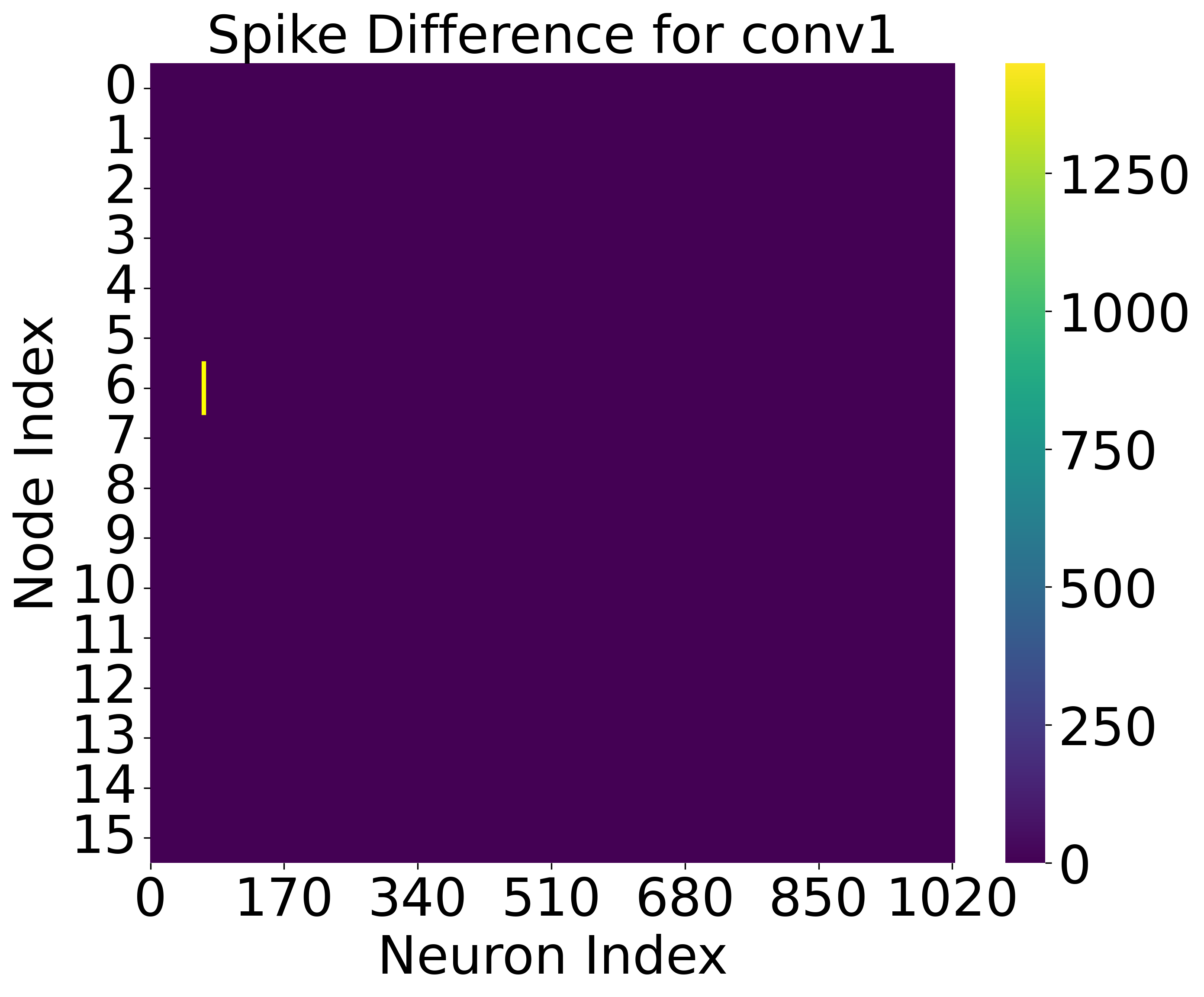}
  \caption{Layer 1}
  \label{fig:sfig1}
\end{subfigure}%
\begin{subfigure}{.25\textwidth}
  \centering
  \includegraphics[width=.95\linewidth]{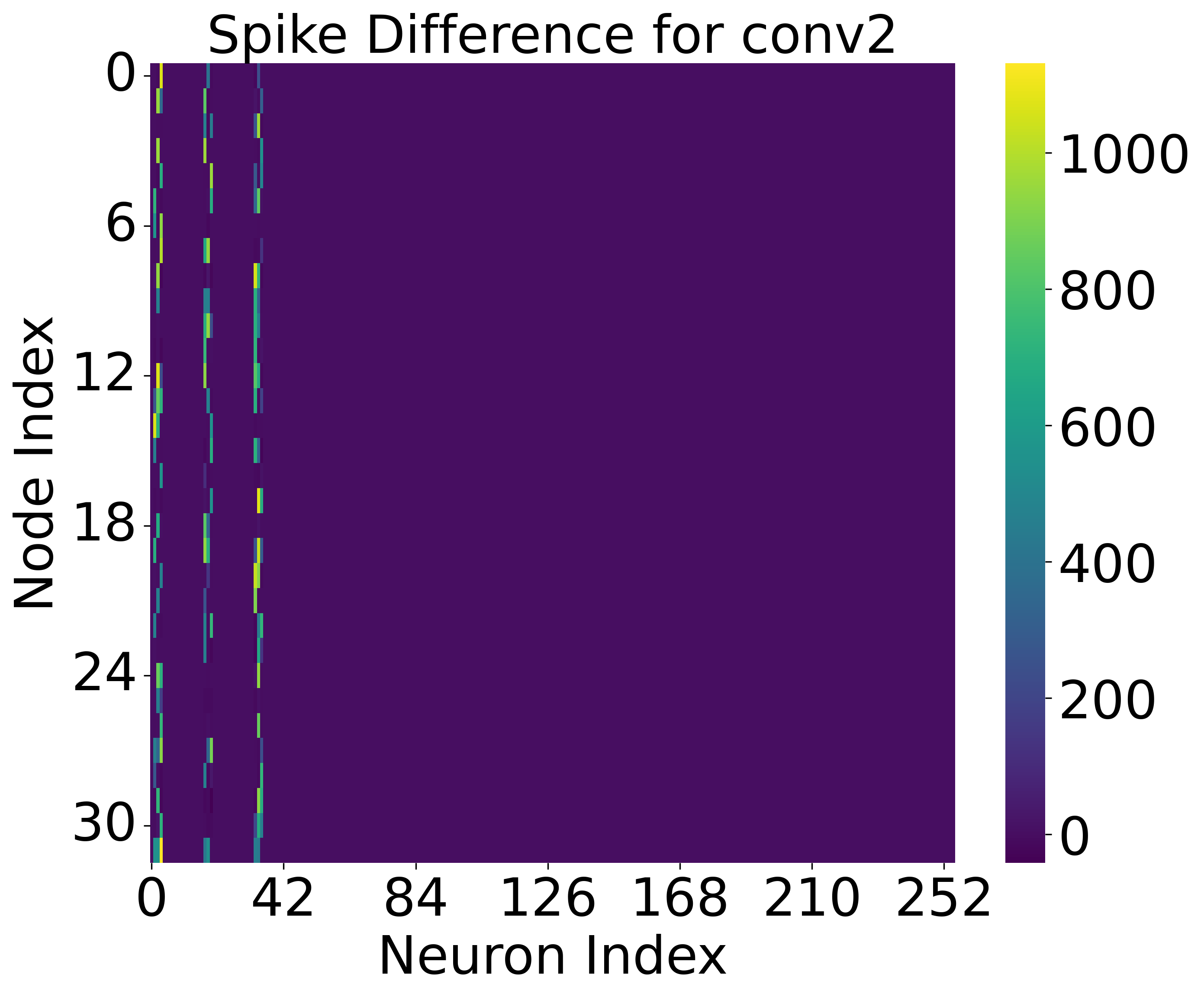}
  \caption{Layer 2}
  \label{fig:sfig2}
\end{subfigure}
\begin{subfigure}{0.25\textwidth}
  \centering
  \includegraphics[width=0.95\linewidth]{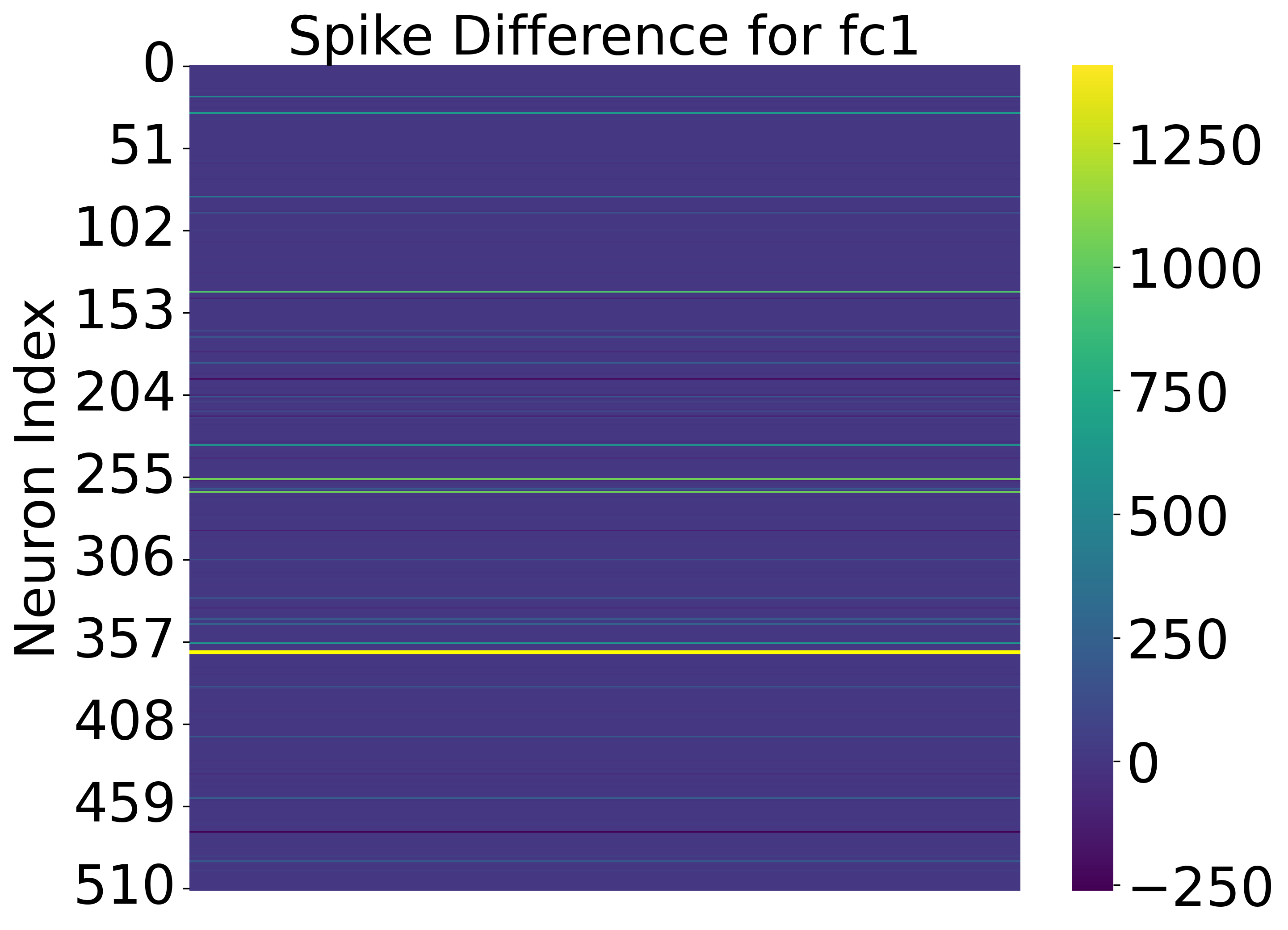}
  \caption{Layer 3}
  \label{fig:sfig3}
\end{subfigure}%
\begin{subfigure}{.25\textwidth}
  \centering
  \includegraphics[width=.95\linewidth]{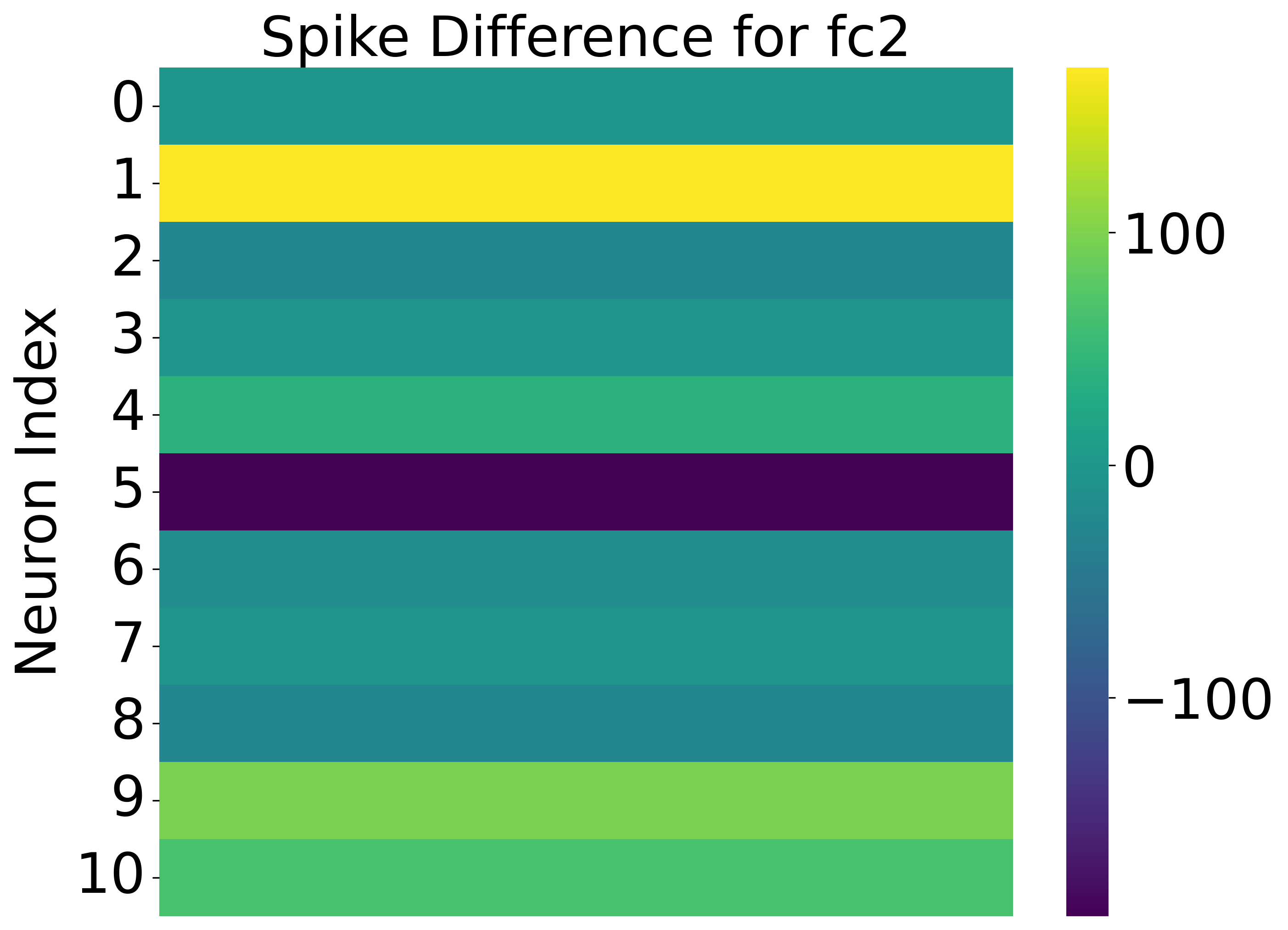}
  \caption{Layer 4}
  \label{fig:sfig4}
\end{subfigure}
\caption{Spike count difference before and after the HT activation for the same dataset sample.}\vspace{-0.4cm}
\label{fig:ibm_spike_difference}
\end{figure}

As another illustration, Fig. \ref{fig:ibm_spike_difference} shows for each neuron in each layer the spike count difference across the complete 1.45 s duration of the input sample before and after the HT activation. The Trojan neuron saturation is easily recognizable in layer 1. At layer 4, we observe that after HT activation the neuron corresponding to class 5 (left arm clockwise gesture), which was the winning class during nominal operation, gives no spikes (high negative spike count), while the neuron corresponding to class 1 (right hand wave gesture), gives more than extra 100 spikes (high positive spike count), leading to a misleading decision about which input hand gesture has occurred.

\section{Hardware Implementations}\label{sec:HT_design}

Herein, we present the hardware implementations of the HT trigger and payload mechanisms. For the payload, we distinguish between analog and digital SNN hardware designs, while the trigger is common for both design paradigms. For the analog design paradigm, we consider an analog spiking neuron at transistor level and we show that the payload mechanism is simply a switch that cuts off one transistor. This approach can be easily adapted for any analog spiking neuron design \cite{IL-BHSE-CDLDHRSCAHFSS-GWWB11}. For the digital design paradigm, we consider the design in \cite{Camunas18} and insert the payload mechanism into a path that is commonly found in all practical hardware platforms for neuromorphic computing \cite{SFMRWQ22}, making it a generic approach virtually applicable to all these architectures. We make a full demonstration of the HT operation on FPGA and we measure the area and power footprint of the HT.

\subsection{HT Trigger mechanism}\label{sec:spike_pattern_checker}

\begin{figure}[t]
    \centering  \includegraphics[width=0.7\columnwidth]{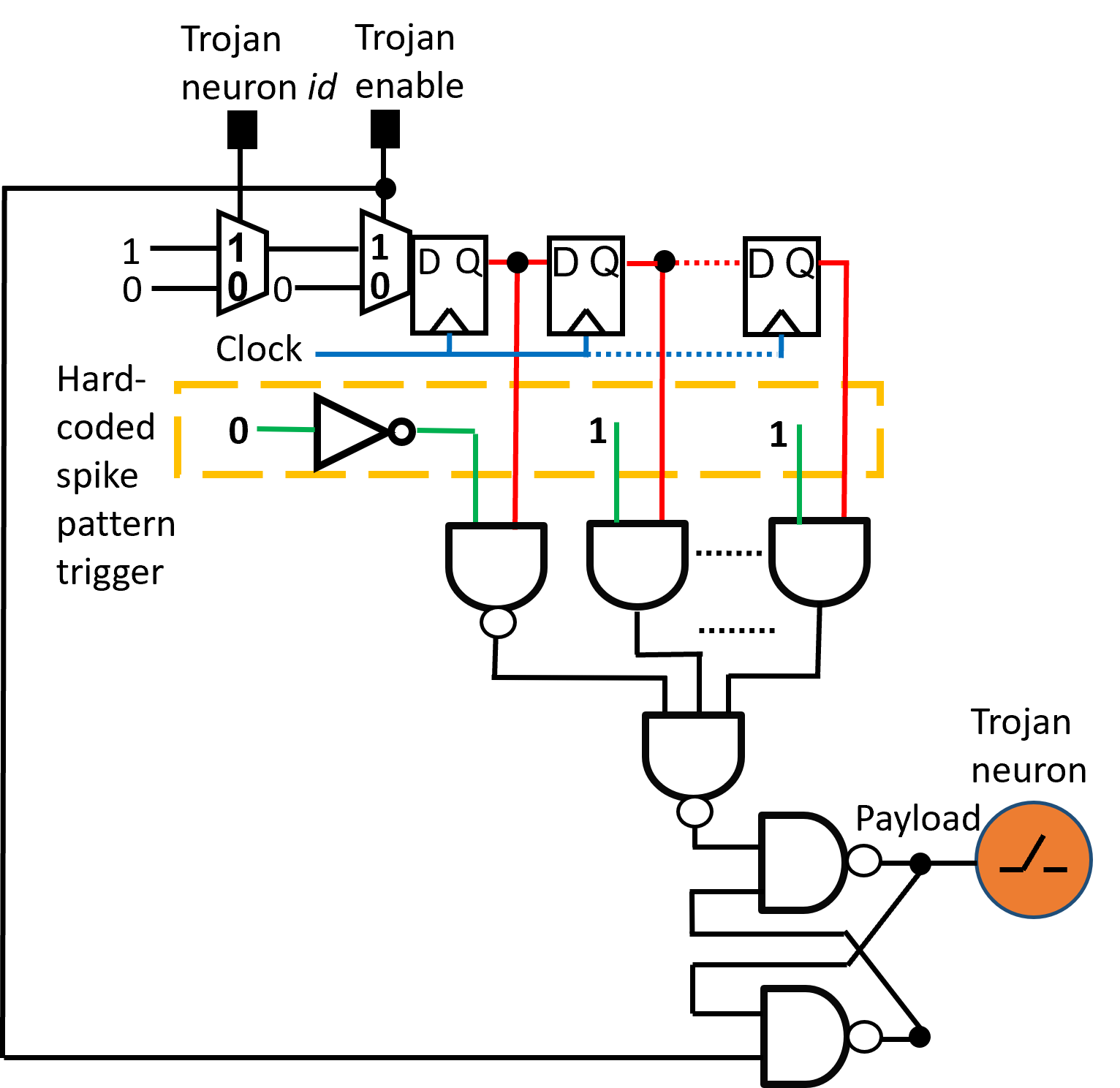}
    \caption{HT trigger mechanism.}
    \label{fig:hardware_Trojan_trigger}\vspace{-0.2cm}
\end{figure}

The trigger mechanism is essentially a spike pattern checker. It takes as input the hard-coded spike pattern trigger and the Trojan neuron's output spike train and, when there is a match between the two, it generates a \texttt{1}-bit trigger signal that delivers the payload back to the Trojan neuron. Fig. \ref{fig:hardware_Trojan_trigger} shows a possible design compatible with the Address-Event Representation (AER) protocol \cite{Bo00}, which is a widely adopted standard in SNN hardware design. In the AER protocol, spikes are discrete events represented by the address $(neuron\_id, t)$ of the emitting neuron. A neuron that emits a spike, rather than sending the actual spike waveform to the target neurons, it sends its address to a router which is then responsible for forwarding the spike to the target neurons. Essentially AER replaces physical connections between neurons with virtual connections, thus enabling a denser integration of neurons, reduced data transmission given the spike sparsity, and reduced power consumption. In Fig. \ref{fig:hardware_Trojan_trigger}, the Trojan neuron is represented by its address $id$. When it emits a spike, a \texttt{1} is shifted into the serial register, otherwise a \texttt{0} is shifted. At any moment, the register contains the Trojan neuron's running spike output train of length equal to the spike pattern trigger. 
The \texttt{AND} gates perform the checking and an SR-Latch is used to store the trigger activation so as to permanently deliver the payload to the Trojan neuron.

\subsection{HT Payload mechanism for analog designs}\label{sec:analog_neuron}

\begin{figure}[t]
    \centering  \includegraphics[width=1.0\columnwidth]{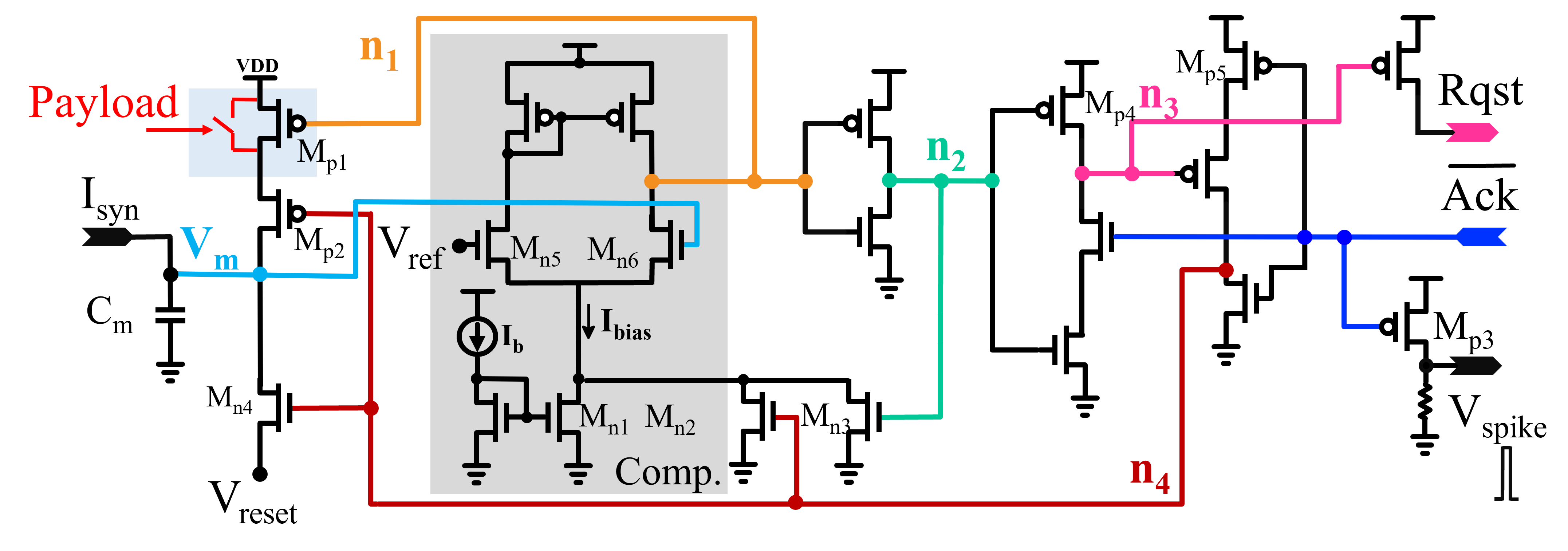}
    \caption{HT Payload implementation for an analog spiking neuron.}\vspace{-0.3cm}
    \label{fig:analog_Trojan_neuron}
\end{figure}

\begin{figure}[t]
    \centering  \includegraphics[width=1.0\columnwidth]{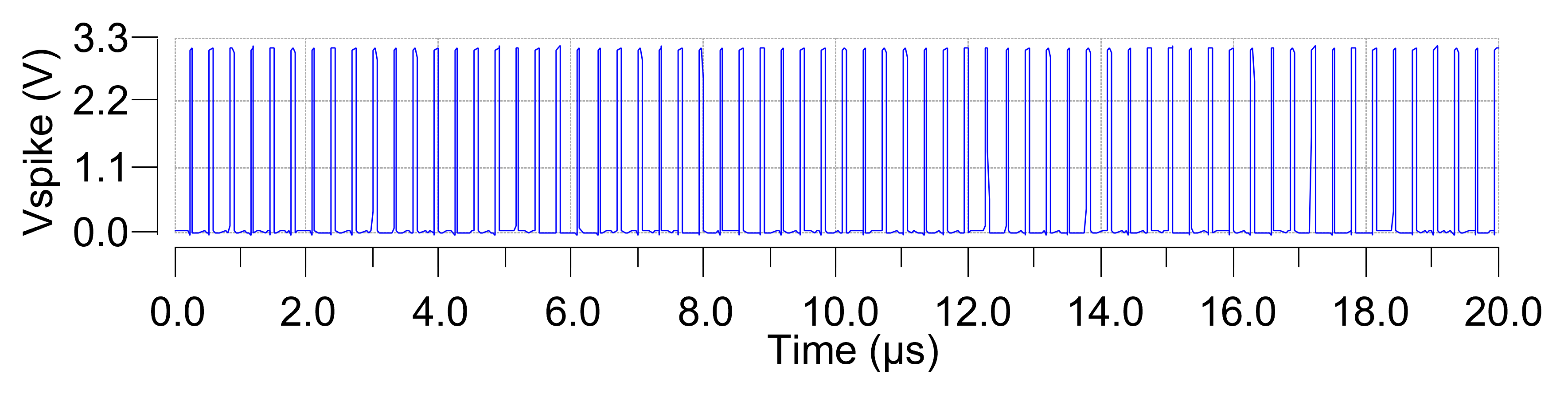}\vspace{-0.2cm}
    \caption{Neuron saturation after HT triggering.}\vspace{-0.4cm}
    \label{fig:neuron_saturation}
\end{figure}

Fig. \ref{fig:analog_Trojan_neuron} shows the transistor-level design of an analog Integrate \& Fire (I\&F) spiking neuron adapted from \cite{serr2006}. 
The neuron takes as input incoming spikes in current mode $I_{syn}$, integrates them on capacitor $C_m$, and when the capacitor voltage $V_m$ reaches a certain threshold $V_{ref}$, the neuron sends a spike request signal \textit{Rqst} to the AER block. The AER block acknowledges back the request with signal $Ack$ which resets the neuron so that it can fire again.

During the charging time of the capacitor, transistors $M_{p1}$ and $M_{n4}$ are off, and transistor $M_{p2}$ is on. As $V_{m}$ increases towards $V_{ref}$, the output of the comparator $n_1$ starts changing state and switches on two transistors: (i) $M_{p1}$, which slowly introduces a positive feedback current that accelerates the charging of the capacitor; and (ii) $M_{n3}$ through node $n_2$, which offers a brief surge in the comparator bias current. These actions combined speed up the transition time of the comparator output. Once the transition is complete, i.e., node $n_{1}$ is low and node $n_2$ is high, node $n_3$ goes low and a spike request signal is sent to the AER block by pulling up line \textit{Rqst}. When the acknowledgment is received, $\overline{Ack}$ pulls up node $n_{4}$ which has three main effects on the neuron circuit:
\textbf{(i)} it turns transistor $M_{n2}$ on to keep the comparator bias current high during the back transitioning; \textbf{(ii)} it turns off transistor $M_{p2}$ which cuts off the positive feedback path to the capacitor; and
\textbf{(iii)} it turns transistor $M_{n4}$ on to reset $V_{m}$ to $V_{reset}$ so that the neuron is able to fire again. Transistor $M_{p3}$ has been added to produce an analog spike waveform when $Ack$ is received.

The HT payload mechanism is simply a switch controlling transistor $M_{p1}$. When the trigger pattern is detected, the switch short-circuits $M_{p1}$ making it permanently stuck-on. This artificially induced defect provokes a constant high feedback current to the capacitor, so the capacitor is always charging even without a current from the incoming synapses. This causes the neuron to saturate as shown by the transistor-level simulation of Fig. \ref{fig:neuron_saturation}. 

\subsection{HT Payload mechanism for digital designs}\label{sec:digital_accelerator}

\begin{figure}[t]
    \centering  \includegraphics[width=0.5\columnwidth]{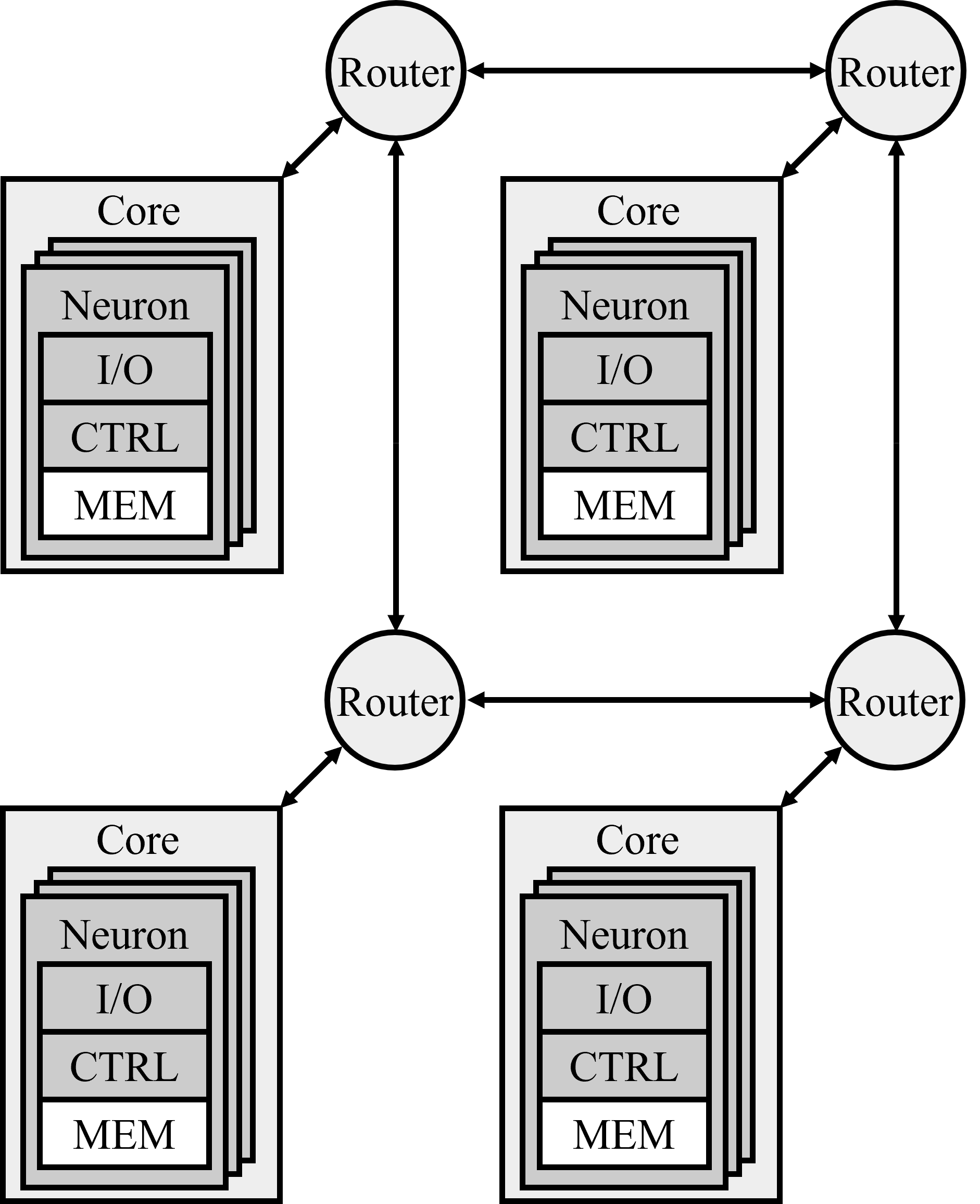}
    \caption{SNN hardware accelerator architecture \cite{SFMRWQ22}.}\vspace{-0.2cm}
    \label{fig:SNN_digital_architecture}
\end{figure}

\begin{figure}[t]
    \centering  \includegraphics[width=1\columnwidth]{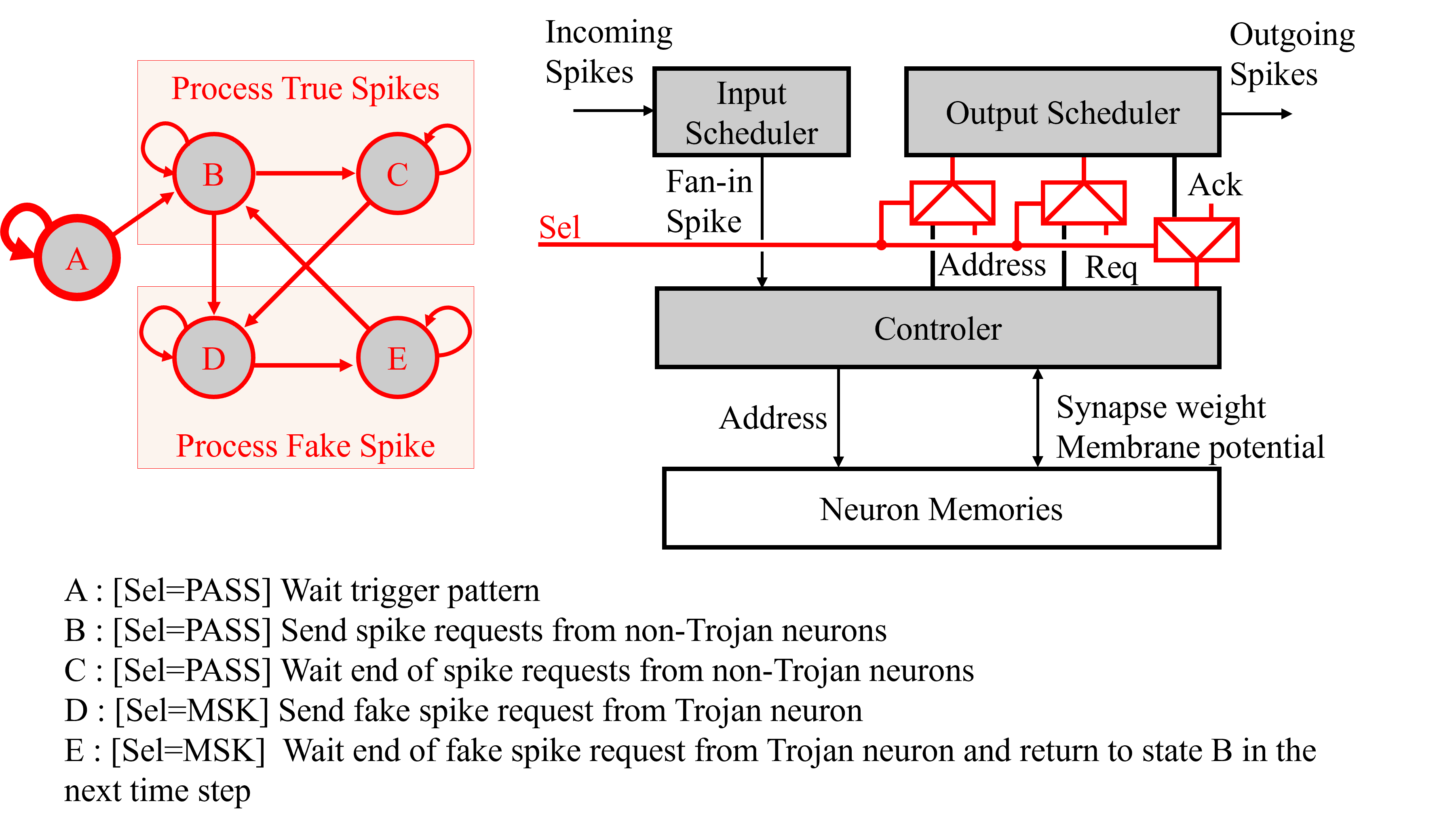}
    \caption{Neuromorphic core architecture embedding the HT payload.}\vspace{-0.2cm}
    \label{fig:core_payload_mechanism}
\end{figure}

\noindent\textbf{Node design.} Modern SNN hardware accelerator designs adopt the architecture shown in Fig. \ref{fig:SNN_digital_architecture} \cite{SFMRWQ22}. Groups of processing elements, i.e., neurons, and data storage, i.e., neuron parameters and synaptic weights, are co-located into cores. This aligns to the near-memory computing principle reducing the overhead of memory access latency and energy consumption. A network-on-chip (NoC) architecture is used to interconnect the many cores and route the spike events based on the AER protocol. 

Fig. \ref{fig:core_payload_mechanism} shows a minimal, generic, block-level schematic of the neuromorphic core, along with the HT payload mechanism shown in red color. In the general case, the core comprises several neurons, for example for implementing a feature map. Incoming spikes from other cores are received by the input scheduler which forwards them to the target neurons in this core based on the spike addresses and network connectivity. At every time step of the internal clock, incoming spikes arriving at this time step are processed sequentially. The neuronal dynamics are implemented into the controller. Specifically, let us consider a source neuron in some other core that sends a spike to a target neuron in this core. When the spike arrives, the controller accesses the memory where the neuron membrane potentials and the synaptic weight are stored. The synaptic weight connecting the source and target neurons is retrieved by the controller which uses it to modulate the incoming spike. Then, the controller calculates the spike's contribution to the target neuron's membrane potential and updates it, checking if it exceeds the threshold. In this case, the neuron emits a spike which is managed by the output scheduler and the controller resets the membrane potential to its resting state so that the neuron can fire again. The output scheduler needs only the address that represents the spike so that it forwards the spike to the right destination neurons according to the network connectivity. It sends a spike request \textit{Rqst} to the AER block and then sends back to the controller an acknowledgment signal \textit{Ack} to process the next spike.

\noindent\textbf{Payload design.} The HT payload mechanism is placed between the controller and output scheduler, illustrated in red color in Fig. \ref{fig:core_payload_mechanism}. It is a generic implementation compatible with practical SNN hardware platforms \cite{SFMRWQ22} and independent of the neuronal dynamics and network connectivity. Specifically, the payload is composed of 3 multiplexers in the paths of the spike address and the $Rqst$ and $Ack$ signals, with the multiplexers being controlled by the \texttt{1}-bit $Sel$ signal whose two states are defined by a Finite State Machine (FSM).

The operation of the HT payload mechanism is as follows. The FSM includes 5 states, namely $A,B,C,D$ and $E$, as shown in Fig. \ref{fig:core_payload_mechanism}. While the Trojan is dormant the FSM is at state $A$ with $Sel=\mbox{PASS}$ which makes the multiplexers transparent, passing the address and signals $Rqst$ and $Ack$ as normal. When the spike pattern trigger appears, the FSM enters state $B$ where at every time step it is checked if any incoming spike has a non-Trojan neuron destination. These true spikes are being processed normally during state $C$, which waits for their processing to be over and then the FSM enters state $D$. In case of no true incoming spikes, the FSM enters state $D$ directly after $B$. During state $D$, the $Sel$ signal changes to $MSK$ mode, masking the normal operation by sending a fake spike emission request from the Trojan neuron. Finally, during state $E$, the FSM waits until the fake spike emission is completed and returns to state B in the next time step. In this way, at every time step the payload mechanism forces the Trojan neuron to emit a fake spike, while the operation of all other neurons in the core is not affected.

\subsection{FPGA Implementation}
\label{sec:fpga_implementation}

\begin{figure}[t]
\centering
\includegraphics[width=0.8\linewidth]{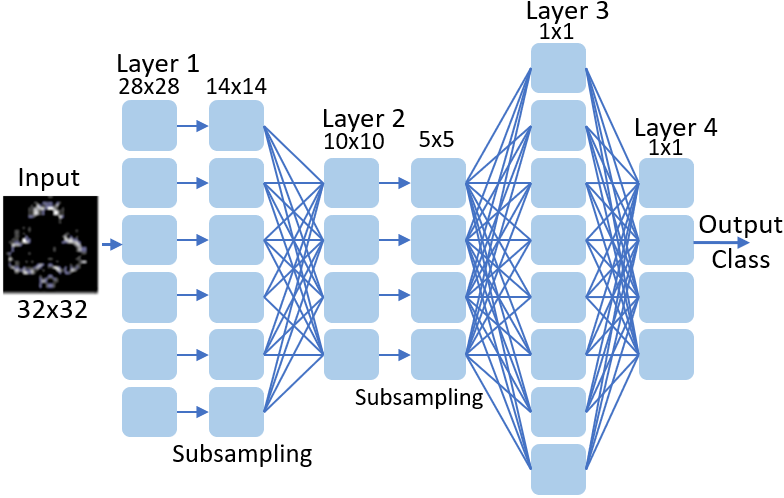}
\caption{SNN model for card symbol recognition.}\vspace{-0.2cm}
\label{fig:card_symbols_SNN}
\end{figure}

The HT trigger and payload mechanisms shown in Figs. \ref{fig:hardware_Trojan_trigger} and \ref{fig:core_payload_mechanism} were designed and integrated into the VHDL code of the SNN hardware accelerator described in \cite{Camunas18}. This SNN hardware accelerator follows the design paradigm shown in Fig. \ref{fig:SNN_digital_architecture} \cite{SFMRWQ22} and the AER spiking communication protocol. The benchmark is card symbol recognition with input acquired from a DVS sensor \cite{Se-GoLi-Ba15}. The SNN model, shown in Fig. \ref{fig:card_symbols_SNN}, is convolutional with four output neurons corresponding to the four card symbols (\texttt{Hearts}, \texttt{Diamonds}, \texttt{Clubs} and \texttt{Spades}). The nominal classification accuracy is 85\%. The Trojan neuron was selected to be neuron $(0,0)$ of the first node in the first layer. The SNN hardware accelerator designs with and without the HT were implemented onto a \textit{Zync UltraScale+ MPSoC ZCU104} FPGA board.

\begin{figure}[t]
\centering
\includegraphics[width=0.8\linewidth]{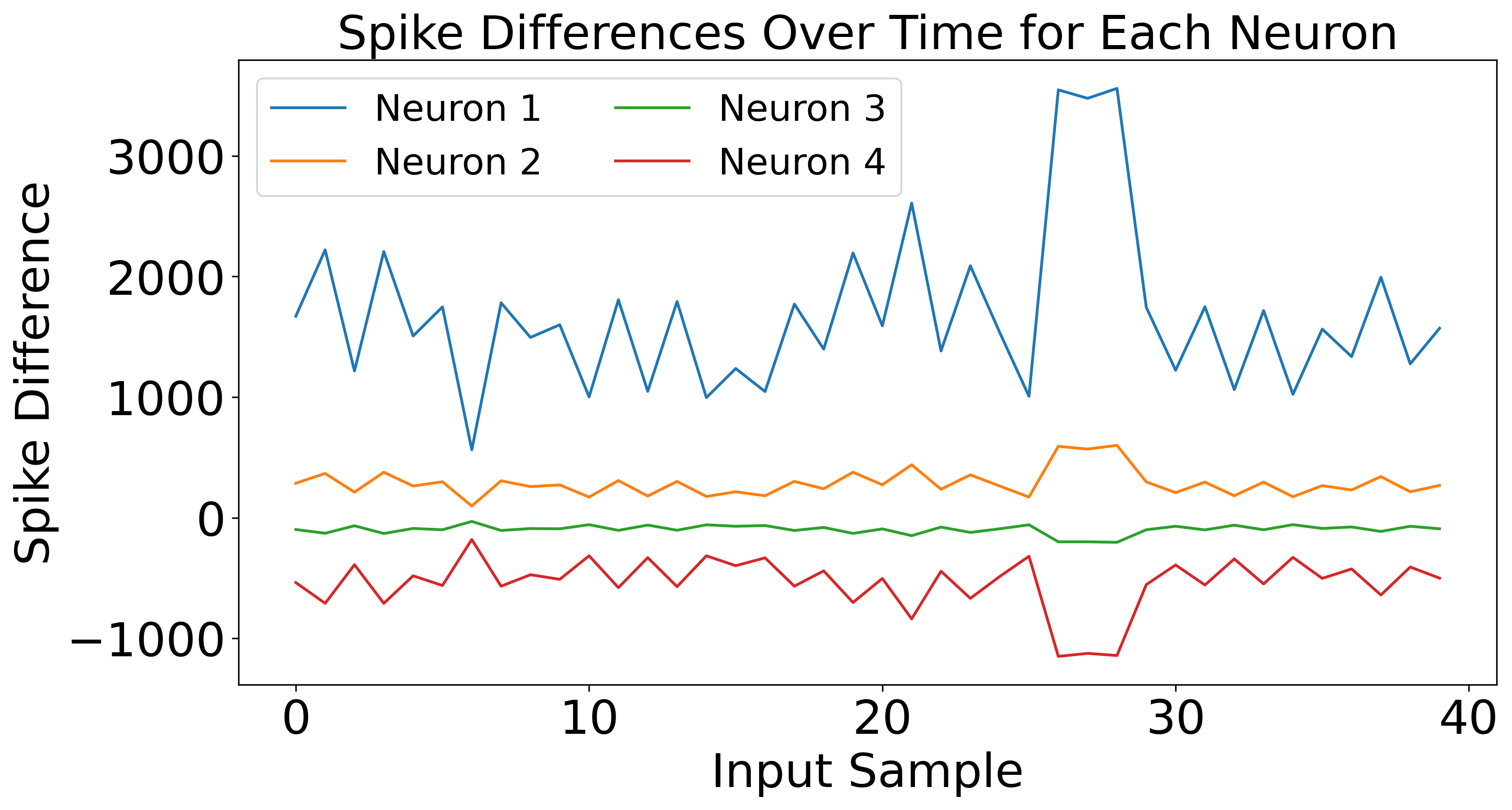}\vspace{-0.1cm}
\caption{HT impact on output spiking neurons measured on the FPGA.}\vspace{-0.0cm}
\label{fig:Trojan_impact_FPGA}
\end{figure}

The HT, once activated, has a catastrophic effect on the classification accuracy which drops from $85\%$ to $25\%$. The output neuron corresponding to \texttt{Hearts} always wins the competition, that is, the network always predicts \texttt{Hearts}. The impact on the output spike trains is illustrated in Fig. \ref{fig:Trojan_impact_FPGA} which shows the per-neuron spike difference between the HT-free and HT-infected designs for 40 input card samples. The Trojan neuron saturation causes output neuron $1$ to spike excessively while output neuron $4$ spikes far less.

\begin{table}[t]
\centering
\scriptsize
\caption{FPGA Resource utilization.}
\begin{tabular}{ | m{2.0cm} | m{1.7cm}| m{2.0cm} | } 
  \hline
   & \textbf{Nominal Design} & \textbf{Trojan-infected Design}\\
  \hline
  \hline
  LUT & $44704$ & $44766$\\ 
  \hline
  LUTRAM & $1404$ & $1404$\\ 
  \hline
  FFs & $45764$ & $45756$\\ 
  \hline
  BRAM & $28$ & $28$\\ 
  \hline
  IO & $8$ & $8$\\ 
  \hline
  BUFG & $3$ & $3$\\ 
  \hline
  Total Power & $3.516W$ & $3.517W$\\ 
  \hline
\end{tabular}
\label{tab:Resource_utilization}\vspace{-0.4cm}
\end{table}

Table \ref{tab:Resource_utilization} shows the footprint of the HT in terms of area overhead and power consumption before its activation. As it can be seen, the HT has minimal impact on the FPGA's resources. A $0.138\%$ increase in LUTs is noticed because of the added FSM. The tiny decrease in FFs is explained by the resource utilization optimization during synthesis. The increase of power consumption due to the continuous operation of the trigger pattern checker is as low as $1$ mW, i.e., a $0.028\%$ power overhead. The results prove that the HT design is indeed very stealthy having a negligible area and power footprint. Note that this overhead is fixed and independent of the size of the SNN model. Thus, as SNN models grow larger, the relative overhead will decrease even further. 

\section{Countermeasures} \label{sec:countermeasures}

According to our threat model, the defender possesses the SNN model, dataset, and SNN hardware accelerator which may be compromised by the HT. However, the defender lacks knowledge of the input trigger and must rely solely on post-silicon detection countermeasures. We distinguish two categories of countermeasures, namely generic and SNN-specific. 

\subsection{Generic countermeasures}

\subsubsection{Reverse engineering}

It involves de-packaging, de-layering and imaging the chip to extract the layout and functionality \cite{FSKWERP17,LWUSEDGRKG19}. Thereafter, the HT can be uncovered with detailed inspection. As SNN hardware accelerators have a modular architecture based on identical nodes, the Trojan neuron may stand out despite the small footprint of the HT mechanism. However, reverse engineering is destructive, time-consuming, and expensive. An adaptive attacker can still bypass this countermeasure using a camouflaging approach by implementing the HT mechanism in all cores to have the same layout, while enabling it exclusively for the core that contains the Trojan neuron.

\subsubsection{Testing}

Logic testing has been proposed to expose HTs in digital designs \cite{CWPPB09,NoFaHe18}. The aim is to develop a dedicated automatic test pattern generation (ATPG) tool to generate test patterns that sensitize suspicious and seldom-activated paths, knowing that HTs are triggered by rare conditions to avoid detection. However, such a tool requires a gate-level hardware model which is not available to the defender according to our threat model. We can envision instead developing an equivalent ATPG tool in the spiking domain implemented in the software framework used to build and train the SNN model. Using this tool, the defender generates a compact set of spatio-temporal inputs to maximize coverage of Trojan activation. This can be done in the defender's premises as it does not involve training. With our algorithm, generating one input to trigger a given Trojan requires the trained model and takes up a few minutes. In our case, the exhaustive number of inputs to guarantee maximum coverage is $d*N*2^d$, where $N$ is the number of neurons and $d$ is the length of the trigger spike pattern, which is unknown to the defender. For example, considering the NMNIST SNN that has $N=1790$ and a trigger spike pattern of $d \leq 10$, the number of inputs is $\approx 18.33*10^6$, which is huge making exhaustive coverage impossible. The defender can reduce this space by performing fault injection to identify critical neurons and focus the analysis only on those \cite{GMCRSC24, SpHaSt24}. Given the large search space, the tool should incorporate statistical methods. Developing an efficient tool can be an area of future research. The generated test inputs by the tool are applied to the SNN accelerator chip and, if it functions after the complete test, then the defender can presume that it is HT-free.

\subsubsection{Statistical side-channel fingerprinting}

Another widely used method for HT detection is statistical side-channel fingerprinting. The idea is to obtain physical chip measurements such that in this measurement space the footprints of HT-infected and HT-free chips are well distinguished \cite{ABKRS07,RaPlTe08,LiBuPa09,DNCB10,KoMi11,NDCPWPRB13,LiHuMa14}. The boundary can be allocated using an one-class classifier trained with the HT-free chip instances. For example, for SNNs, parametric measurements can include power supply transient signals, regional supply currents, timing variations of output spike trains, etc. The aim is that these measurements capture the always-on operation of the trigger pattern checker and the loading of the Trojan neuron due to the addition of the payload mechanism. However, this approach requires a golden chip or at least a trusted hardware model, thus it is not applicable under our threat model. Besides, the minimal footprint of the proposed HT makes it challenging to distinguish from noise and normal variations.

\subsubsection{Side-channel analysis}

There exist other non-destructive side-channel analyses that can detect the HT, e.g., optical circuit analysis \cite{SSWCHP14}, electromagnetic emanation (EM) measurements \cite{SKMH14,NNBGD16, HZGJ17}, thermal map analysis \cite{TLFHC19}, backscattering \cite{NCPZ19}, and laser probing \cite{SMTGFT20}. For these analyses a golden chip model is preferred but not mandatory. They can be effective given that the trigger pattern checker is always-on while neurons fire sparsely.

\subsection{SNN specific counter-measures}

\subsubsection{Neuron monitoring} The neurons' outputs could be monitored for flagging neuron saturation in real-time. After detection, the application can be suspended to prevent misleading decisions. For example, in the context of fault detection, it is proposed to count the number of spikes for groups of neurons at the feature map level, and use these spike counts as features to the input of a classifier which is trained to detect abnormal spiking activity \cite{SpSt23}. However, the necessary hardware provisions to record internal spikes are typically not available in SNN hardware accelerators. Even if this was case, the attacker can easily manipulate the output of the Trojan neuron before it is fed to the spike counter so as to fool the classifier. 

\subsubsection{Input filters}
The input trigger is an out-of-distribution input with respect to typical inputs met during the application, meaning that it differs from normal spatial and temporal spike patterns. To this end, we can envision adding a filter to the input of the SNN that screens out incoming outlier inputs. Such input filters have been proposed for pre-processing the inputs to reduce background noise and improve the overall accuracy of the SNN task \cite{PaBaOr18,L-BP-PMG-RJ-MLD19}, but also to detect adversarial attacks \cite{MPMMS21}. However, these filters are not yet adapted to SNN hardware accelerators. The main challenge is to reduce the percentage of false positives. 



\section{Conclusion} \label{sec:conclusion}

We proposed an input-triggered HT attack in SNNs. The HT trigger and payload mechanisms are condensed around one Trojan neuron. The trigger is a spike pattern at the output of the Trojan neuron delivered by the input trigger message and the payload saturates the Trojan neuron, with the continuous spikes propagating to the SNN output and changing its decision. We proposed a methodology to select the Trojan neuron and an algorithm to generate the input trigger. We also proposed a generic hardware implementation of the HT mechanisms for both analog and digital design paradigms. The attack is shown by simulation on three popular SNN benchmarks and on an FPGA implementation of an SNN hardware accelerator. The FPGA implementation results demonstrated a very stealthy HT of 0.028\% power consumption increase having a tiny area footprint of only 0.138\% LUTs. In terms of future work, we plan to focus on countermeasures, in particular an ATPG tool in the spiking domain and an on-line input filter which seem the most promising defenses.


\bibliographystyle{IEEE}

\end{document}